\documentclass{article}

\usepackage{microtype}
\usepackage{graphicx}
\usepackage{subcaption}
\usepackage{booktabs} % for professional tables

\usepackage{hyperref}

\PassOptionsToPackage{numbers,sort&compress}{natbib}

\usepackage[accepted]{icml2024}

\usepackage{amsmath}
\usepackage{amssymb}
\usepackage{mathtools}
\usepackage{amsthm}

\usepackage{tikz}

\usepackage[capitalize,noabbrev]{cleveref}

\theoremstyle{plain}

\theoremstyle{definition}

\theoremstyle{remark}

\usepackage[textsize=tiny]{todonotes}

\icmltitlerunning{Rethinking Global Average Pooling}

\begin{document}

\twocolumn[
    \icmltitle{\texorpdfstring{Rethinking Global Average Pooling: \\Your Classifier Is Secretly a Multi-Instance Learner}{Rethinking Global Average Pooling: Your Classifier Is Secretly a Multi-Instance Learner}}

    % It is OKAY to include author information, even for blind
    % submissions: the style file will automatically remove it for you
    % unless you've provided the [accepted] option to the icml2024
    % package.

    % List of affiliations: The first argument should be a (short)
    % identifier you will use later to specify author affiliations
    % Academic affiliations should list Department, University, City, Region, Country
    % Industry affiliations should list Company, City, Region, Country

    % You can specify symbols, otherwise they are numbered in order.
    % Ideally, you should not use this facility. Affiliations will be numbered
    % in order of appearance and this is the preferred way.
    % \icmlsetsymbol{equal}{*}

    \begin{icmlauthorlist}
        \icmlauthor{Aray Karjauv}{}
        % \icmlauthor{Firstname2 Lastname2}{equal,yyy,comp}
        % \icmlauthor{Firstname3 Lastname3}{comp}
        % \icmlauthor{Firstname4 Lastname4}{sch}
        % \icmlauthor{Firstname5 Lastname5}{yyy}
        % \icmlauthor{Firstname6 Lastname6}{sch,yyy,comp}
        % \icmlauthor{Firstname7 Lastname7}{comp}
        % %\icmlauthor{}{sch}
        % \icmlauthor{Firstname8 Lastname8}{sch}
        % \icmlauthor{Firstname8 Lastname8}{yyy,comp}
        %\icmlauthor{}{sch}
        %\icmlauthor{}{sch}
    \end{icmlauthorlist}

    % \icmlaffiliation{tu}{Department of XXX, University of YYY, Location, Country}
    % \icmlaffiliation{gtarc}{Company Name, Location, Country}
    % \icmlaffiliation{sch}{School of ZZZ, Institute of WWW, Location, Country}

    % \icmlcorrespondingauthor{Aray Karjauv}{Aray.Karjauv@tu-berlin.de}
    % \icmlcorrespondingauthor{Firstname2 Lastname2}{first2.last2@www.uk}

    % You may provide any keywords that you
    % find helpful for describing your paper; these are used to populate
    % the "keywords" metadata in the PDF but will not be shown in the document
    \icmlkeywords{Self-supervised learning, foundation models, multi-instance learning, explainable AI, CNN, ViT}

    \vskip 0.3in

    \begin{center}
    \begin{minipage}{\textwidth}
        \centering
        \captionsetup{type=figure}
        \captionsetup[subfigure]{singlelinecheck=false,justification=centering,margin={0pt,50pt}}
        \subcaptionbox{ConvNeXt}[0.33\textwidth]{%
            \includegraphics[width=\linewidth]{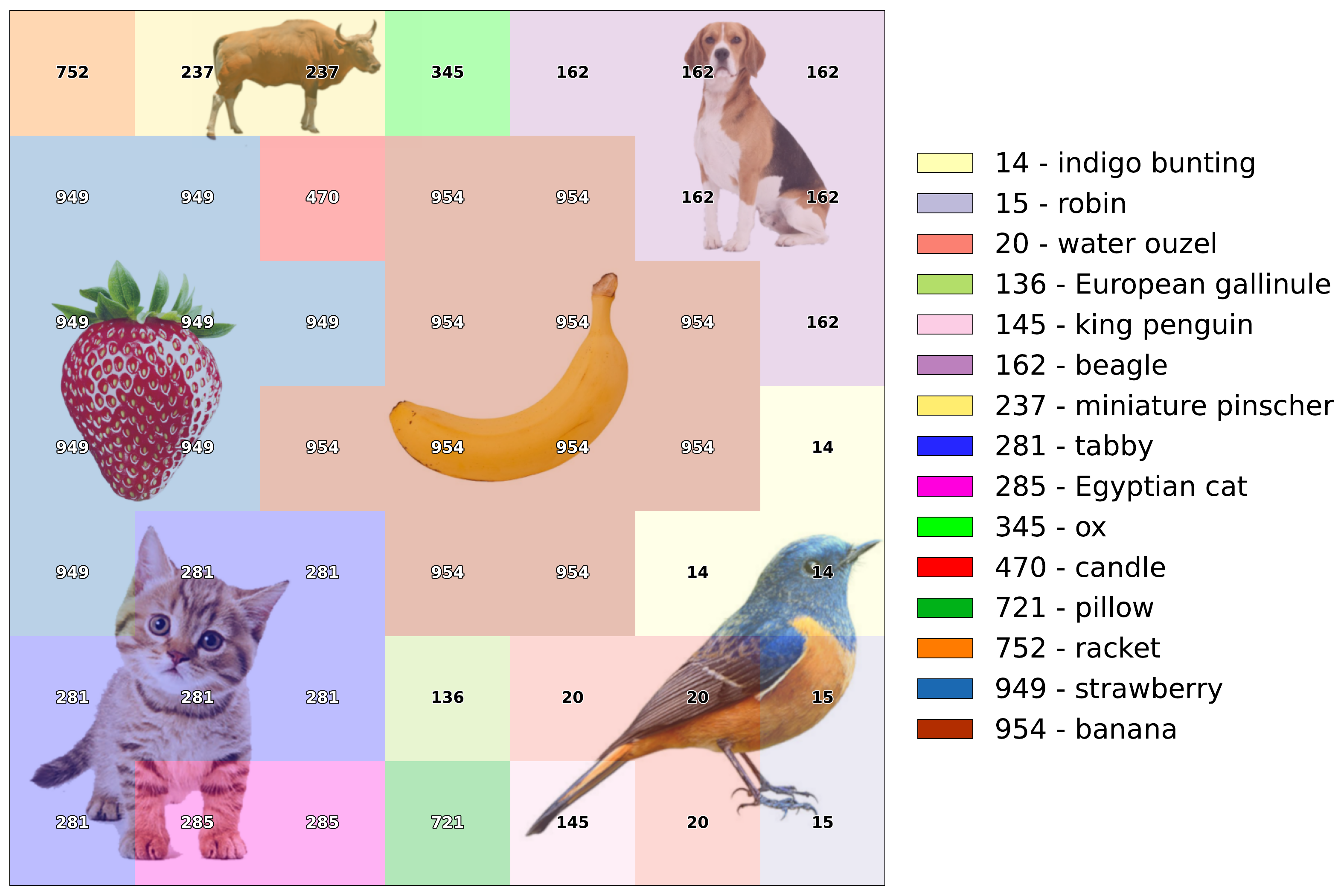}%
        }%
        \subcaptionbox{Swin}[0.33\textwidth]{%
            \includegraphics[width=\linewidth]{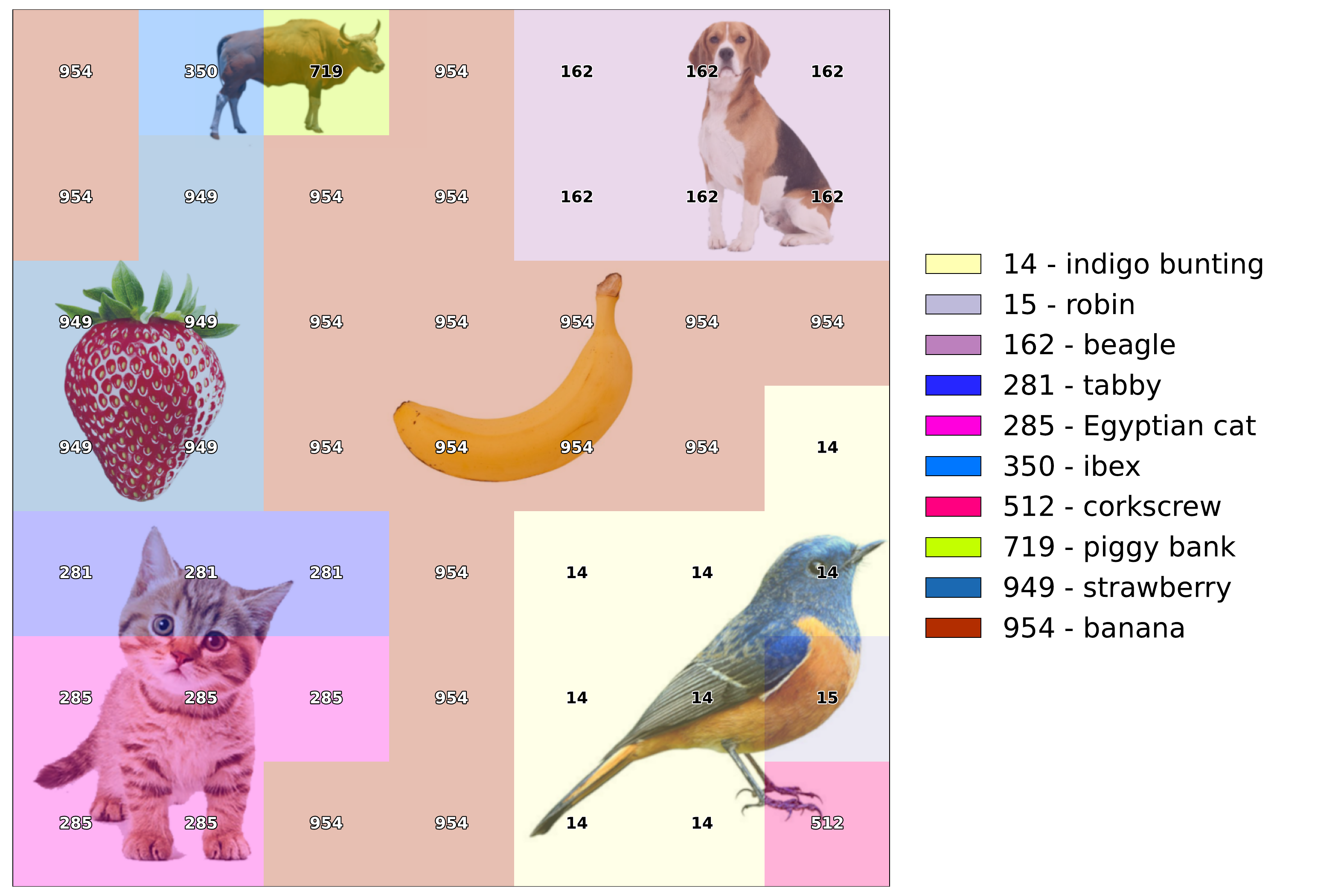}%
        }%
        \subcaptionbox{MaxViT}[0.33\textwidth]{%
            \includegraphics[width=\linewidth]{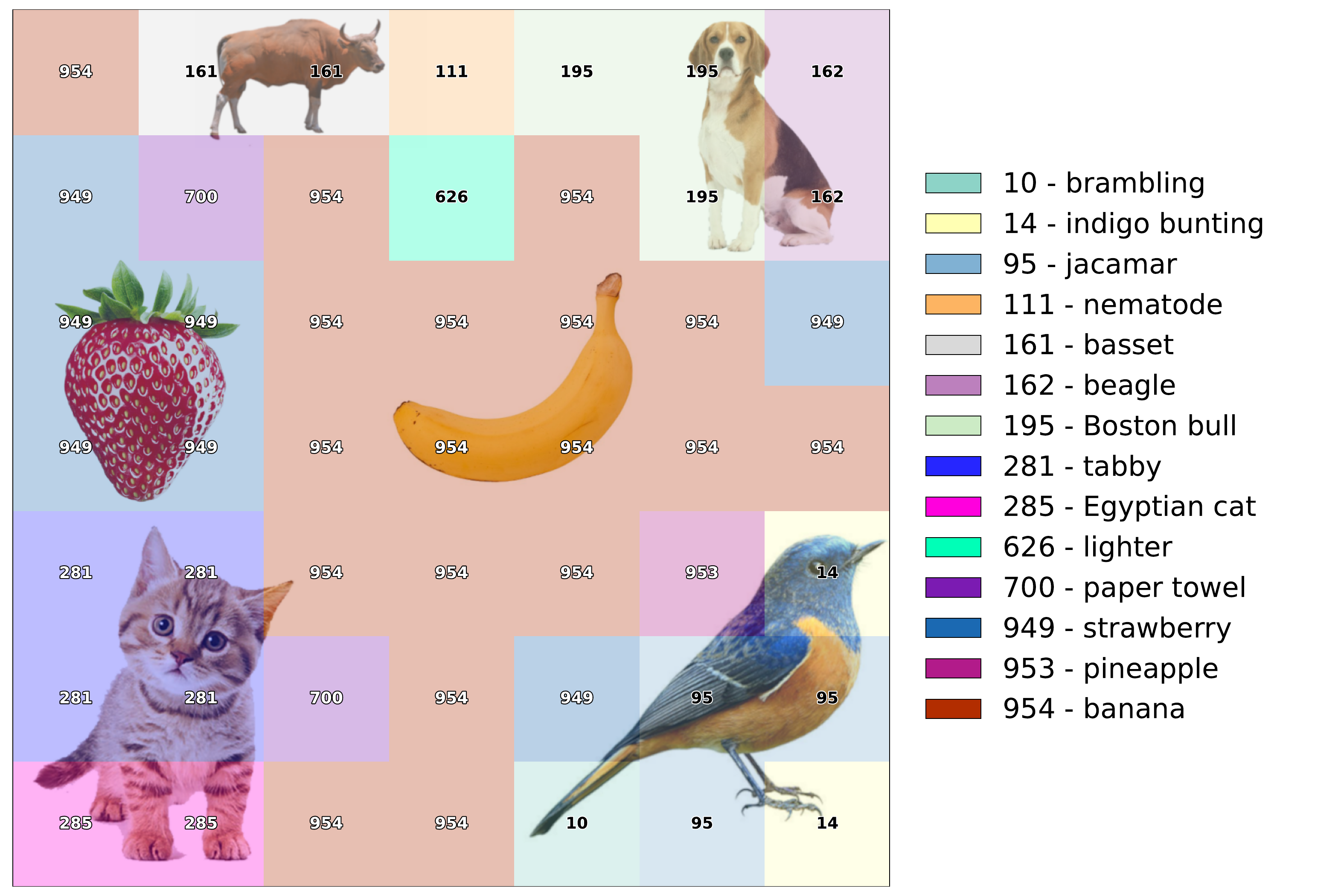}%
        }%
        \addtocounter{figure}{-1}
        \captionof{figure}{Spatial predictions from three pretrained ImageNet classifiers trained with GAP and image-level labels only. Each grid cell reports the top-1 class obtained by applying the trained classification head to the final feature grid before pooling. The resulting maps reveal coarse localization signals for multiple objects that are otherwise averaged into a single image-level prediction.}
        \label{fig:spatial-predictions}
    \end{minipage}
    \vskip 0.2in
\end{center}

]

% this must go after the closing bracket ] following \twocolumn[ ...

% This command actually creates the footnote in the first column
% listing the affiliations and the copyright notice.
% The command takes one argument, which is text to display at the start of the footnote.
% The \icmlEqualContribution command is standard text for equal contribution.
% Remove it (just {}) if you do not need this facility.

% \printAffiliationsAndNotice{}  % leave blank if no need to mention equal contribution
% \printAffiliationsAndNotice{\icmlEqualContribution} % otherwise use the standard text.

\begin{abstract}
    Modern image classifiers widely adopt global average pooling (GAP) followed by a linear classification head. This linearity ensures that the image-level logits equal the average of logits obtained by applying the classification head pointwise to the feature grid prior to GAP. Consequently, standard classifiers may inherently retain spatial class evidence that remains recoverable even when the image-level prediction is incorrect. This structure naturally suggests a multiple-instance learning (MIL) interpretation, where an image is viewed as a bag of spatial instances. Within this formulation, we demonstrate that standard classifiers trained with a single label per image can still learn the intended classification task in multi-object scenes. We further exploit this property to decompose image-level logits into a prediction grid, providing a post-hoc diagnostic to extract spatial class evidence that GAP otherwise obscures. Our systematic evaluation reveals that off-the-shelf models consistently recover the ground-truth class within foreground regions. The MIL interpretation further suggests that common classifier failures reflect known limitations of mean aggregation.
\end{abstract}

\def\arraystretch{1.3}%

\section{Introduction}

Natural images rarely contain a single isolated object. An image labeled \texttt{garden spider} may also include a web, leaves, and background textures. Yet standard image classification models collapse this complex scene into a single image-level decision. The widespread adoption of top-5 accuracy evaluation tacitly acknowledges this limitation~\citep{russakovsky2015imagenet}. Top-5 evaluation permits multiple valid predictions per image, but it remains a global metric and cannot localize the spatial evidence. A classifier might assign high probabilities to both \texttt{garden spider} and \texttt{spider web}, but image-level metrics cannot distinguish whether these scores arise from distinct physical objects, surrounding context, or entangled features.

Reflecting this limitation, the community has devoted substantial effort to re-annotating existing datasets with multi-object labels~\cite{beyer2020we,yun2021relabeling,gao2022luss}. These efforts expose the incompleteness of single-label evaluation but still treat the model output as an opaque image-level prediction. In this setting, multi-label metrics can indicate which categories a model predicts but cannot determine whether a decision stems directly from the physical object or from supporting contextual cues. We therefore explore whether spatial class evidence can be recovered from the internal representations without additional supervision or retraining.

\subsection{Background}

Convolutional neural networks (CNNs) process an image through a hierarchy of convolutional layers, producing spatial activation maps at each stage. Each activation map contains $C$-dimensional feature vectors arranged over an $H \times W$ spatial grid, where $C$ denotes the number of channels. As depth increases, spatial resolution typically decreases while feature channels become increasingly selective for higher-level semantic patterns~\cite{zeiler2014visualizing}. The final activation map therefore provides a coarse spatial grid of high-level visual evidence, where each spatial feature vector encodes semantic information from a corresponding region of the input image.

The spatial structure of these representations has long motivated visual explanation methods, which seek to expose where a classifier finds evidence for its predictions. Class activation mapping (CAM) and related gradient-based methods weight activation maps by classifier weights or gradients to produce class-specific heatmaps~\cite{zhou2016learning,selvaraju2017grad}. These methods demonstrate that activation maps in CNNs contain localized semantic evidence, but they remain tied to a queried class and are typically used to explain a single image-level prediction. More recently, a simple average across the channel dimension has been shown to localize learned concepts without relying on class conditioning, indicating that activation maps retain spatially structured semantic evidence~\cite{karjauv2024lafam}. A similar interpretation extends to vision transformers (ViTs)~\cite{dosovitskiy2020image}, where the image is divided into patches and each patch is embedded as a spatial token. Through self-attention blocks, patch tokens exchange information across the image, enabling early aggregation of global information~\cite{raghu2021do}. The final token grid can therefore be viewed as a high-level spatial representation analogous to the final activation map in CNNs~\cite{zhou2022extract}. We use the term spatial feature grid as an architecture-neutral description of both representations.

Modern image classifiers~\cite{liu2022convnet,liu2021swin,tu2022maxvit} widely use global average pooling (GAP)~\cite{lin2013network} before classification. This aggregation substantially reduces the number of classifier parameters and supports translation invariance by making predictions less sensitive to the absolute location of feature responses. For a final spatial feature grid with $C$ channels and spatial dimensions $H \times W$, GAP produces a single $C$-dimensional image-level feature vector, which is typically passed to a linear classification head. This aggregation structure naturally suggests a multiple-instance learning (MIL)~\cite{dietterich1997solving,maron1998framework} interpretation of GAP-based classification models.

MIL defines each labeled example as a bag of instances, where supervision is provided only for the bag as a whole and individual instance labels are unobserved. Under this formulation, an image can be viewed as a bag of spatial instances corresponding to the feature vectors in the final feature grid. When the classification head is linear, the image-level logits can be decomposed into the average of local logits obtained by applying the same linear head pointwise to the final feature grid, as formalized in Eq.~\eqref{eq:gap_mil}. This dense readout yields a coarse spatial class-score map whose entries represent local class evidence before global aggregation. Once training is complete, removing the bottleneck imposed by GAP exposes spatial class evidence that emerges implicitly from image-level supervision alone, as shown in Fig.~\ref{fig:spatial-predictions}.

This reframing shifts the analysis from image-level correctness to the distribution of spatial class evidence. Under a linear head after GAP, a failure of the image-level prediction does not necessarily imply that target-class evidence is absent from the feature grid. Strong responses from a small object can be averaged with weak or competing responses elsewhere, reducing their effect on the final logit. Dense readouts make this failure mode directly inspectable by exposing the local logits. This inference-time effect is distinct from the broader MIL problem of credit assignment. In the MIL view, image-level labels supervise only the bag prediction and leave the relevant spatial instances unspecified. This connects GAP-based classifiers to the choice of aggregation rule in MIL and suggests that their failure modes may reflect known limitations of mean aggregation.

Within this framework, we evaluate the dense readout as a post-hoc diagnostic for trained classifiers across convolutional and transformer architectures\footnote{\url{https://github.com/karray/revising_gap}.}. On ImageNet-1K, bounding-box annotations and foreground occlusion allow local predictions to be attributed to objects and context, and the recovered classes are consistent with multi-label human reannotations~\cite{beyer2020we}. On ImageNet-A~\cite{hendrycks2021natural}, where image-level accuracy collapses, the target class nonetheless remains recoverable within the spatial grid. On MS-COCO, linear heads trained only on pooled embeddings of frozen self-supervised encoders recover foreground object evidence when applied before pooling, extending the diagnostic beyond end-to-end supervised training. Complementing these inference-time analyses, we study the learning problem under controlled conditions on a synthetic dataset where each image contains multiple valid object instances but receives only a single label during training. Models trained from random initialization learn the classification task in these multi-object scenes, providing direct evidence that GAP-based classifiers can learn from bags of spatial instances under single-label supervision. Together, these results connect failure modes in GAP-based classifiers to known limitations of mean-aggregation MIL and make pooling itself an object of analysis.
\section{Related Work}

Early weakly supervised object localization (WSOL) work showed that CNN classifiers trained only with image-level labels can coarsely localize objects~\cite{oquab2015localization}, although evaluating such localization reliably is known to be nontrivial~\cite{choe2020wsol}. CAM and its successors made this spatial evidence explicit by weighting each activation map according to its contribution to a target class score, yielding class-discriminative heatmaps~\cite{zhou2016learning,selvaraju2017grad,chattopadhay2018grad}. The same formulation extends to GAP-based ViTs, whose patch tokens preserve spatial correspondence to image regions (see Appendix~\ref{app:vit-cam}). However, these methods require a target class to be specified in advance, either manually or from the model prediction. The resulting map is therefore constrained to evidence for the queried class, even when the image contains multiple semantically meaningful objects. In contrast, the dense readout yields a single class-score map covering all classes.

Prior work shows that local visual cues can support strong image classification, even when their spatial arrangement is largely ignored~\cite{brendel2019approximating}. This motivates treating spatial features as local evidence vectors, especially in natural images where multiple objects or contextual cues may coexist. Several WSOL works have recognized a related connection by framing image classification as a MIL problem, where image-level labels supervise latent spatial instances. \citet{oquab2015localization} propose max pooling in the final layer to capture the most discriminative spatial instance, while weakly supervised segmentation methods similarly cast pixels or regions as latent instances supervised only by image-level labels~\cite{pinheiro2015image,pathak2015fully}. Later approaches adopt this MIL view for weakly supervised model design~\cite{kraus2016classifying,durand2016weldon,durand2017wildcat,ilse2018attention,araujo2024key}, primarily focusing on modifying training objectives, architectures, or aggregation rules. We instead study what this perspective implies for standard image classifiers, analyzing their spatial evidence and failure modes as consequences of instance aggregation.

Interpreting GAP-based classifiers through a MIL lens provides a complementary way to examine well-known limitations of image classification, which prior work has often studied through additional annotations, revised evaluation protocols, or targeted benchmark redesign~\cite{beyer2020we,yun2021relabeling,vasudevan2022does,idrissi2022imagenet,recht2019imagenet,shankar2020evaluating,tsipras2020imagenet,xiao2020noise,geirhos2020shortcut,hendrycks2021natural}. \citet{beyer2020we} re-annotate ImageNet validation images with multi-label annotations, showing that the single label is often an incomplete description of image content. \citet{hendrycks2021natural} further demonstrate that natural adversarial examples can still defeat standard image-level classifiers. We argue that standard classifiers already contain a spatial signal that can be used to diagnose such failures without additional annotation or retraining. In our setting, existing annotations serve only as evaluation ground truth.

\section{Method}\label{sec:method}

Let $x$ be an input image and let $E(\cdot)$ denote an encoder producing a spatial tensor
\begin{equation}
F = E(x) \in \mathbb{R}^{C \times H \times W},
\end{equation}
where $C$ is the channel dimension and $(H, W)$ are the spatial dimensions.

We denote by $h_{u,v} \in \mathbb{R}^{C}$ the spatial feature vector at coordinate $(u,v)$, corresponding to a location in the last feature map for CNNs or to a patch token on the final token grid for ViTs. Flattening the spatial grid gives $N = HW$ spatial feature vectors, which we index as $\{h_i\}_{i=1}^{N}$.

GAP produces a single embedding
\begin{equation}
\bar{h} = \frac{1}{N}\sum_{i=1}^{N} h_i,
\end{equation}
which is passed through a classification head $f:\mathbb{R}^{C}\to\mathbb{R}^{K}$ to obtain class logits $z=f(\bar{h})$ for $K$ classes.

When the head is a single fully connected layer, $f(h)=\Theta^\top h+b$ with $\Theta\in\mathbb{R}^{C\times K}$ and $b\in\mathbb{R}^{K}$, applying the same head to each spatial feature vector gives
\begin{equation}
Z_{:,u,v} = \Theta^\top h_{u,v}+b,
\qquad
Z \in \mathbb{R}^{K \times H \times W},
\end{equation}
where $b$ is shared across spatial locations. The image-level logits are then equal to the spatial average of these local logits:
\begin{equation}
z = f(\bar{h})
= \frac{1}{N}\sum_{u=1}^{H}\sum_{v=1}^{W} Z_{:,u,v}.
\label{eq:gap_mil}
\end{equation}
We refer to $Z$ as the \emph{dense readout} and to its entries as \emph{spatial class scores}.

A GAP-based classifier is therefore mathematically equivalent to a mean-aggregation MIL model, in which the image is a bag of instances $\{h_i\}$ and the bag logit is the mean of instance-level logits.

However, treating spatial feature vectors as independent instances in a bag requires an additional assumption. Strict independence does not hold, since CNNs inherently aggregate local information through overlapping receptive fields, while ViTs globally redistribute context across spatial tokens through self-attention. We nonetheless adopt it as a working assumption, as is common in MIL formulations over image patches.

\section{Experiments}
\label{sec:experiments}

To demonstrate that GAP-based classifiers retain spatial class evidence, we conduct three experiments. On ImageNet, bounding-box annotations and controlled occlusions test whether class-score maps align with foreground objects and context. On MS-COCO~\citep{lin2014microsoft}, we test whether the same approach extends to pretrained self-supervised backbones kept frozen, with only linear classification heads trained on spatial features using single positive image-level labels. We additionally train GAP-based classifiers from scratch on a synthetic multi-object dataset. In this setting, each image contains several valid objects, but supervision is provided for only one of them, allowing us to isolate whether spatial class evidence can emerge from deliberately incomplete single-label supervision. Dataset construction, sampling rules, architecture details, and optimization settings are provided in Appendix~\ref{app:experimental-details}.

\subsection{Evaluation Protocol}
\label{sec:eval-protocol}

Evaluation is performed on the class-score map $Z \in \mathbb{R}^{K \times G \times G}$, where $G$ denotes the model-specific grid resolution. Foreground annotations are projected onto this grid, and cells overlapping any annotated foreground region are marked as foreground.

We construct three inputs that are forwarded independently through the model. Alongside the unmodified image, we create a \emph{w/o BG} variant that occludes all pixels outside the foreground region and a \emph{w/o FG} variant that occludes all pixels inside the foreground region.

For ImageNet and MS-COCO, we report grid-level and image-level metrics. \emph{Loc.}\ is top-1 classification accuracy on foreground grid cells. \emph{FG Det.}\ is the fraction of images where the evaluation class is predicted in at least one foreground grid cell. \emph{BG Act.}\ is the fraction of images where the evaluation class is predicted in at least one background grid cell. \emph{Loc.\ (w/o BG)} recomputes foreground localization after background occlusion and tests whether foreground evidence alone suffices for correct spatial prediction. \emph{BG Act.\ (w/o FG)} recomputes background activation after foreground occlusion and measures how often context alone triggers the evaluation class. \emph{Top-1 Acc.}\ and \emph{Top-5 Acc.}\ are standard image-level accuracies. For ImageNet, we additionally report \emph{ReaL Acc.}\ and \emph{ReaL Rec.}, defined below.

\subsection{ImageNet Bounding Box Evaluation}
\label{sec:exp-imagenet}

\paragraph{Models.}
We evaluate five pretrained ImageNet-1K classifiers instantiated from \texttt{timm}~\citep{wightman2019timm}. Architectural details are reported in Table~\ref{tab:imagenet-models} of Appendix~\ref{app:experimental-details}.

\paragraph{Dataset.}
We use the ILSVRC localization task annotations, which provide bounding boxes for the ImageNet-1K validation set~\cite{russakovsky2015imagenet}. Each image belongs to a single synset, so all bounding boxes within an image share the same class label. Multiple boxes per image correspond to multiple instances of the same class. We additionally evaluate the same classifiers on ImageNet-A~\citep{hendrycks2021natural}, a natural-adversarial ImageNet benchmark. Since ImageNet-A does not provide bounding-box annotations, we report only image-level accuracy and spatial detection, denoted \emph{Det.}, which measures whether the target class is predicted in at least one grid cell.

\paragraph{Protocol specifics.}
The foreground region is the union of ILSVRC bounding boxes filled at pixel resolution and then projected to the model-specific spatial grid. The ImageNet ground truth is used as the evaluation class. We additionally report \emph{ReaL Acc.}, which counts the image-level top-1 prediction as correct if it matches any ReaL label~\citep{beyer2020we}, and \emph{ReaL Rec.}, the fraction of ReaL labels recovered by the set of classes predicted anywhere on the grid.

\subsection{MS-COCO Evaluation}
\label{sec:exp-coco}

\paragraph{Dataset.}
MS-COCO contains natural scenes with multiple objects from 80 categories and per-instance segmentation masks~\citep{lin2014microsoft}. As with ILSVRC bounding boxes, segmentation masks are projected onto the model-specific class-score grid. For training, we sample one class uniformly from the set of classes present in each image and use it as the only positive image-level label.

\paragraph{Models.}
We evaluate frozen pretrained backbones from three families: contrastive language-image pretraining (CLIP~\citep{radford2021learning}), self-distillation (DINOv1~\citep{caron2021emerging}, DINOv2~\citep{oquab2023dinov2}, DINOv3~\citep{simeoni2025dinov3}), and masked image modeling (MAE~\citep{he2022masked}, BEiT-3~\citep{wang2023image}). All models are instantiated from \texttt{timm}~\citep{wightman2019timm} and detailed in Table~\ref{tab:coco-models} of Appendix~\ref{app:experimental-details}. For each backbone, we train a linear classification head on image-level embeddings, independently of the aggregation used during pretraining, using cross-entropy on the sampled single-label targets while keeping the backbone frozen. Features are normalized to unit $\ell_2$ norm and passed through dropout with $p=0.3$ before the linear projection. At evaluation, the same head is applied independently at every spatial position to produce class-score maps, following the ImageNet evaluation protocol.

\subsection{Synthetic Single-Label Supervision with Multiple Objects}
\label{sec:exp-synmil}

To isolate the effect of incomplete image-level supervision, we construct a synthetic multi-object dataset of simple geometric figures. Each image contains 3--5 objects sampled without replacement from 20 geometric symbol classes, with randomized position, scale, rotation, color, rendering order, and background. Although several objects in an image may correspond to valid semantic labels, only one visible object is selected as the image-level target label during training, while the remaining objects are treated as unlabeled co-occurring positives. We train \emph{ResNet-18} and \emph{ViT-S/32} from random initialization under this supervision regime and evaluate whether their spatial class scores assign the correct classes to the corresponding visible objects. Full dataset construction and evaluation details are given in Appendix~\ref{app:synmil-details}.

\section{Results}

Table~\ref{tab:bbox-eval} reports performance on the ImageNet-1K validation set. The foreground detection rate exceeds 90\% across all evaluated architectures. Classifiers therefore successfully recognize the target object locally, and top-5 accuracy closely tracks this high local detection rate. Spatial class scores also recover valid ReaL labels that GAP suppresses, causing \emph{ReaL Rec.} to consistently exceed \emph{ReaL Acc.} by 1.2 to 7.1 absolute points.

\begin{table}[t]
\centering
\setlength{\tabcolsep}{1.4pt} % Shrinks the space between columns
\scriptsize % Drops the font size down one more notch to fit the data
\caption{%
Evaluation on ImageNet-1K with ILSVRC boxes projected to the spatial grid. \emph{Loc.}\ is top-1 classification accuracy on foreground grid cells. \emph{FG Det.}\ and \emph{BG Act.}\ measure whether the ground truth is predicted in at least one foreground or background grid cell, respectively. \emph{(w/o BG)} and \emph{(w/o FG)} denote occluded variants. \emph{Top-1 Acc.}, \emph{Top-5 Acc.}, and \emph{ReaL Acc.}\ are baseline image-level accuracies. \emph{ReaL Rec.}\ is the fraction of ReaL labels recovered by classes predicted anywhere on the grid. All values are percentages.%
}
\label{tab:bbox-eval}
\smallskip
\begin{tabular}{@{}l ccc cc cccc @{}}
\toprule
Model & \shortstack{Top-1\\Acc.} & \shortstack{Top-5\\Acc.} & \shortstack{FG Det.} & \shortstack{Loc.} & \shortstack{Loc.\\(w/o BG)} & \shortstack{BG Act.} & \shortstack{BG Act.\\(w/o FG)} & \shortstack{ReaL\\Acc.} & \shortstack{ReaL\\Rec.} \\
\midrule
EfficientNet  & 78.4 & 94.3 & 93.7 & 44.0 & 42.2 & 55.3 & 24.6 & 84.7 & 91.8 \\
ResNet-50     & 80.3 & 95.1 & 94.3 & 33.7 & 33.5 & 34.5 & 28.7 & 86.1 & 92.0 \\
ConvNeXt      & 83.7 & 96.7 & 94.7 & 58.3 & 57.0 & 55.2 & 34.7 & 88.3 & 90.9 \\
Swin          & 83.1 & 96.4 & 94.2 & 75.2 & 76.0 & 89.2 & 35.8 & 87.7 & 89.8 \\
MaxViT        & 84.7 & 96.9 & 94.8 & 77.3 & 77.1 & 90.3 & 37.9 & 88.5 & 89.7 \\
\bottomrule
\end{tabular}%
\end{table}

Architectural differences are especially clear in background activation dynamics under occlusion (Table~\ref{tab:bbox-eval}). On unmodified images, ViT-based models predict the target class in background regions (\emph{BG Act.}) at a high rate (89.2\% and 90.3\%). Foreground occlusion (\emph{BG Act.\ (w/o FG)}) reduces this rate to 35.8\% and 37.9\%. This steep drop suggests that background regions receive target-class evidence from the visible foreground object, consistent with the global context mixing induced by ViT self-attention~\cite{raghu2021do}. Once the object is removed, the source evidence vanishes, and the target-class logits in the background drop. Conversely, CNNs exhibit a substantially lower baseline \emph{BG Act.}\ (34.5\% to 55.3\%) that decreases less sharply under occlusion. For a specific subset of classes, we observe an anomalous inversion where CNN target-class predictions in the background increase when the foreground is occluded (see Appendix~\ref{app:context}).

\begin{table}[t]
\centering
\setlength{\tabcolsep}{3pt} % Shrinks the space between columns
\scriptsize % Drops the font size down one more notch to fit the data
\caption{%
Top-1 and Top-5 evaluation on ImageNet-A. \emph{Acc} is standard image-level classification accuracy, and \emph{Det.} is spatial detection. All values are percentages.
}\label{tab:imagenet-a}
\smallskip
  \begin{tabular*}{\columnwidth}{@{\extracolsep{\fill}} l c c c c @{}}
\toprule
& \multicolumn{2}{c}{Top-1} & \multicolumn{2}{c}{Top-5} \\
\cmidrule(lr){2-3} \cmidrule(lr){4-5}
\textbf{Model} & \textbf{Acc} & \textbf{Det.} & \textbf{Acc} & \textbf{Det.} \\
\midrule
EfficientNet & \ 3.8 & 30.6 & 16.6 & 57.5 \\
ResNet-50    & \ 7.2 & 38.9 & 24.8 & 66.9 \\
ConvNeXt     & 17.9 & 51.9 & 43.9 & 78.8 \\
\midrule
Swin         & 18.0 & 51.1 & 42.9 & 76.1 \\
MaxViT       & 23.3 & 56.9 & 49.2 & 79.5 \\
\bottomrule
\end{tabular*}
\end{table}

Table~\ref{tab:imagenet-a} evaluates model robustness on ImageNet-A. Standard image-level accuracy is severely degraded, ranging from 3.8\% (EfficientNet) to 23.3\% (MaxViT). However, evaluating the spatial class scores reveals much higher \emph{Det.} This suggests that many image-level failures on ImageNet-A arise during spatial aggregation while localized class evidence persists in the feature grid.

Table~\ref{tab:coco_metrics} reports MS-COCO results for linear classification heads trained on aggregated embeddings from frozen self-supervised backbones and then applied to feature grids before GAP to obtain spatial logit maps at validation time. Consistent with the ImageNet results, these spatial logits yield high foreground detection rates (\emph{FG Det.}), peaking at 93.5\% for DINOv2.

\begin{figure}[t]
  \centering
  \begin{subfigure}[t]{0.273\textwidth}
    \centering
    \includegraphics[width=\linewidth]{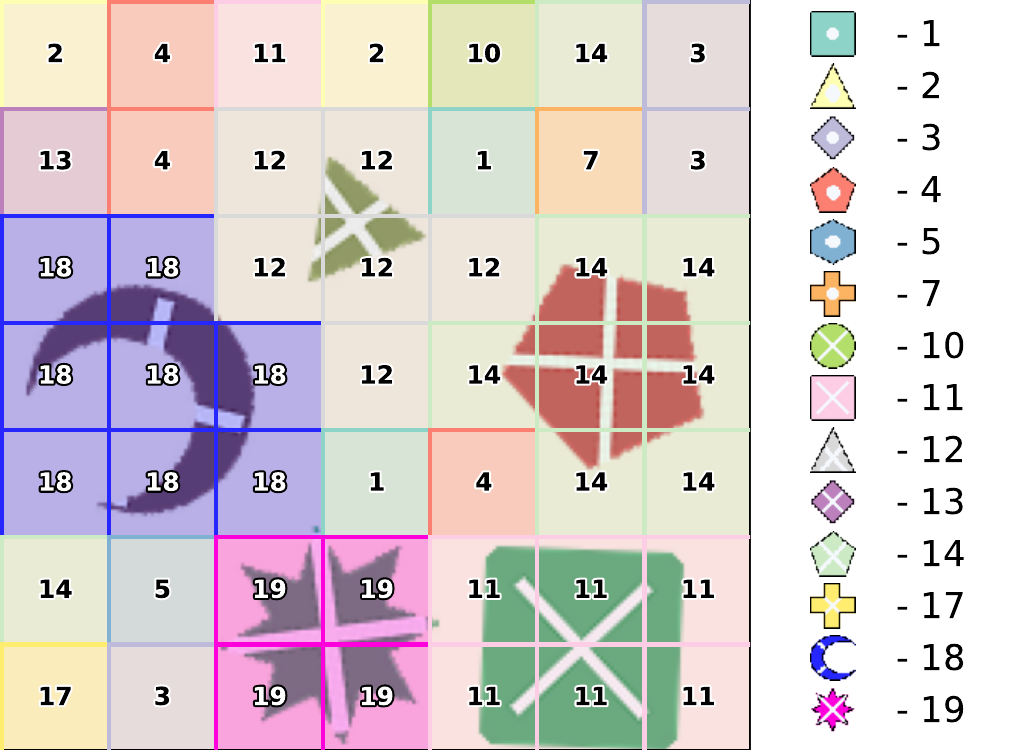}
    \caption{ResNet-18}
  \end{subfigure}
  \hfill
  \begin{subfigure}[t]{0.20\textwidth}
    \centering
    \includegraphics[width=\linewidth]{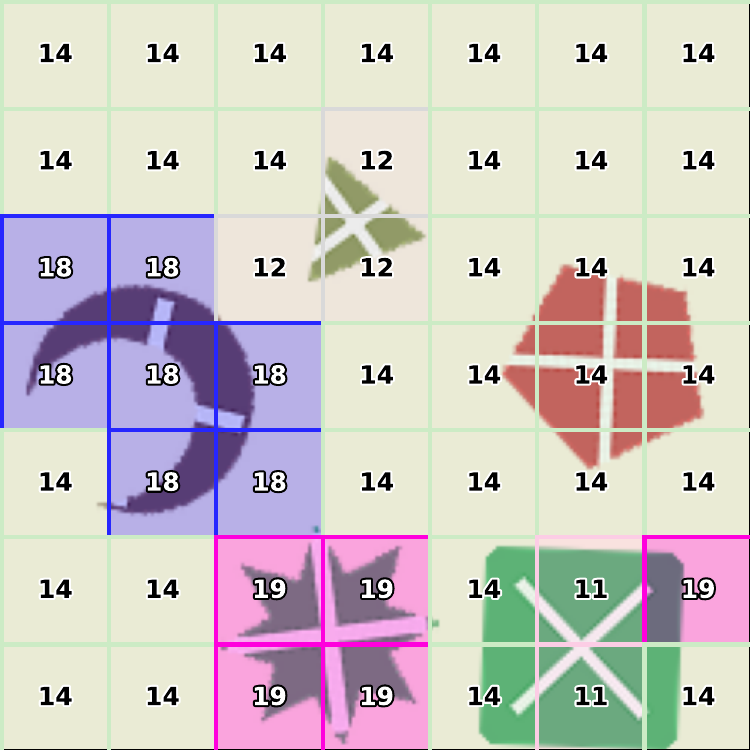}
    \caption{ViT-S/32}
  \end{subfigure}

  \caption{Qualitative synthetic multi-object evaluation. Each panel overlays the input image with predicted class IDs from the dense readout, obtained by applying the classification head to each spatial feature vector before GAP.}
  \label{fig:synth}
\end{figure}

To isolate whether spatial class evidence can emerge under incomplete single-label supervision, we evaluate GAP-based models trained from random initialization on the synthetic multi-object dataset. Top-1 accuracy is measured against the sampled image-level target, whereas localization accuracy (\emph{Loc.}) measures whether foreground cells are assigned any correct visible object class. Image-level accuracy remains near 26\% for both models. This matches the expected rate for a model that predicts one of the $n \in \{3,4,5\}$ visible objects and is scored against the uniformly sampled target, $\mathbb{E}[1/n] \approx 26.1\%$, indicating that both models optimize the ambiguous image-level objective comparably well. In contrast, spatial localization accuracy is substantially higher, reaching 94\% for ResNet-18 and 63\% for ViT-S/32 (Fig.~\ref{fig:synth}). The lower ViT-S/32 value should be interpreted in the context of the lightweight training recipe used for this synthetic experiment, as detailed in Appendix~\ref{app:synmil-details}.

\begin{table}[t]
\centering
\setlength{\tabcolsep}{3pt} % Shrinks the space between columns
\scriptsize % Drops the font size down one more notch to fit the data
\caption{%
Evaluation on the MS-COCO validation set using segmentation annotations. \emph{Top-1 Acc.}\ and \emph{Top-5 Acc.}\ are the image-level baseline classification accuracies. \emph{Loc.}\ is the classification accuracy at spatial positions overlapping any ground-truth segmentation mask. \emph{FG Det.}\ is the fraction of images with at least one correct prediction inside the segmentation mask. \emph{BG Act.}\ is the fraction of images where the object class is predicted at least once outside the segmentation mask. \emph{BG Act.\ (w/o FG)} denotes the background activation metric when the foreground region is occluded. All values are percentages.%
}
\label{tab:coco_metrics}
\smallskip
\begin{tabular*}{\columnwidth}{@{\extracolsep{\fill}} l c c c c c c @{}}
\toprule
Model & \shortstack{Top-1\\Acc.} & \shortstack{Top-5\\Acc.} & \shortstack{FG Det.} & Loc. & \shortstack{BG Act.} & \shortstack{BG Act.\\(w/o FG)} \\
\midrule
CLIPResNet-50 & 47.1 & 88.8 & 50.7 & 44.8 & 75.7 & 53.1 \\
CLIPConvNeXt & 47.8 & 90.4 & 89.4 & 59.4 & 89.1 & 82.0 \\
CLIPViT-B/16 & 46.5 & 89.0 & 74.5 & 61.8 & 88.8 & 73.3 \\
DINOv1ResNet-50 & 45.9 & 86.2 & 87.1 & 59.0 & 83.8 & 77.4 \\
DINOv1ViT-B/16 & 44.2 & 83.1 & 51.6 & 16.1 & 90.5 & 84.2 \\
DINOv2ViT-B/14 & 47.7 & 91.6 & 93.5 & 69.0 & 93.2 & 77.1 \\
DINOv3ViT-B/16 & 47.7 & 91.5 & 92.1 & 77.9 & 91.1 & 75.4 \\
DINOv3ConvNeXt & 47.5 & 90.0 & 90.3 & 66.3 & 81.4 & 76.5 \\
MAEViT-B/16 & 40.2 & 77.5 & 78.4 & 54.7 & 81.8 & 73.8 \\
BEiT-3 ViT-B/16 & 43.0 & 81.6 & 80.5 & 60.6 & 84.8 & 73.4 \\
\bottomrule
\end{tabular*}
\end{table}

\section{Discussion}

Applying the classification head pointwise to spatial feature vectors before global pooling exposes a coarse spatial readout in standard image-level classifiers. Across ImageNet, ImageNet-A, and MS-COCO, the results suggest that these readouts often reveal class evidence that is weakened or suppressed by global aggregation. On ImageNet, \emph{ReaL Rec.}\ exceeds \emph{ReaL Acc.}, indicating that valid labels are frequently recovered somewhere in the spatial grid. On ImageNet-A, spatial detection remains substantially higher than image-level accuracy, suggesting that localized target evidence can persist under natural adversarial distribution shifts. On MS-COCO, frozen self-supervised backbones also exhibit localized foreground class evidence when probed with a linear head trained solely under single-label supervision on image-level embeddings. This shows that applying the trained head before aggregation can expose spatial diagnostic signals that are otherwise collapsed into an averaged image-level logit vector. Finally, the synthetic experiment isolates this effect in a controlled setting, showing that object localization can arise from single-label supervision even in multi-object scenes.

Dense readouts also extend image-level analyses of class co-occurrence, competition, and label uncertainty. Prior work has shown that ImageNet contains images for which multiple labels may be semantically valid, either because several objects co-occur or because the visual evidence supports closely related categories~\cite{beyer2020we,yun2021relabeling}. Such analyses remain limited to image-level predictions. Spatial class scores expose the regions from which competing class evidence originates, allowing class competition to be analyzed with respect to objects, object parts, and contextual structures. This can help disentangle competition between object instances from competition driven by shared attributes such as fur, texture, shape, or background context.

\begin{figure*}[th]
  \centering
  \begin{subfigure}[b]{0.245\textwidth}
    \centering
    \includegraphics[width=\linewidth, height=4cm, keepaspectratio]{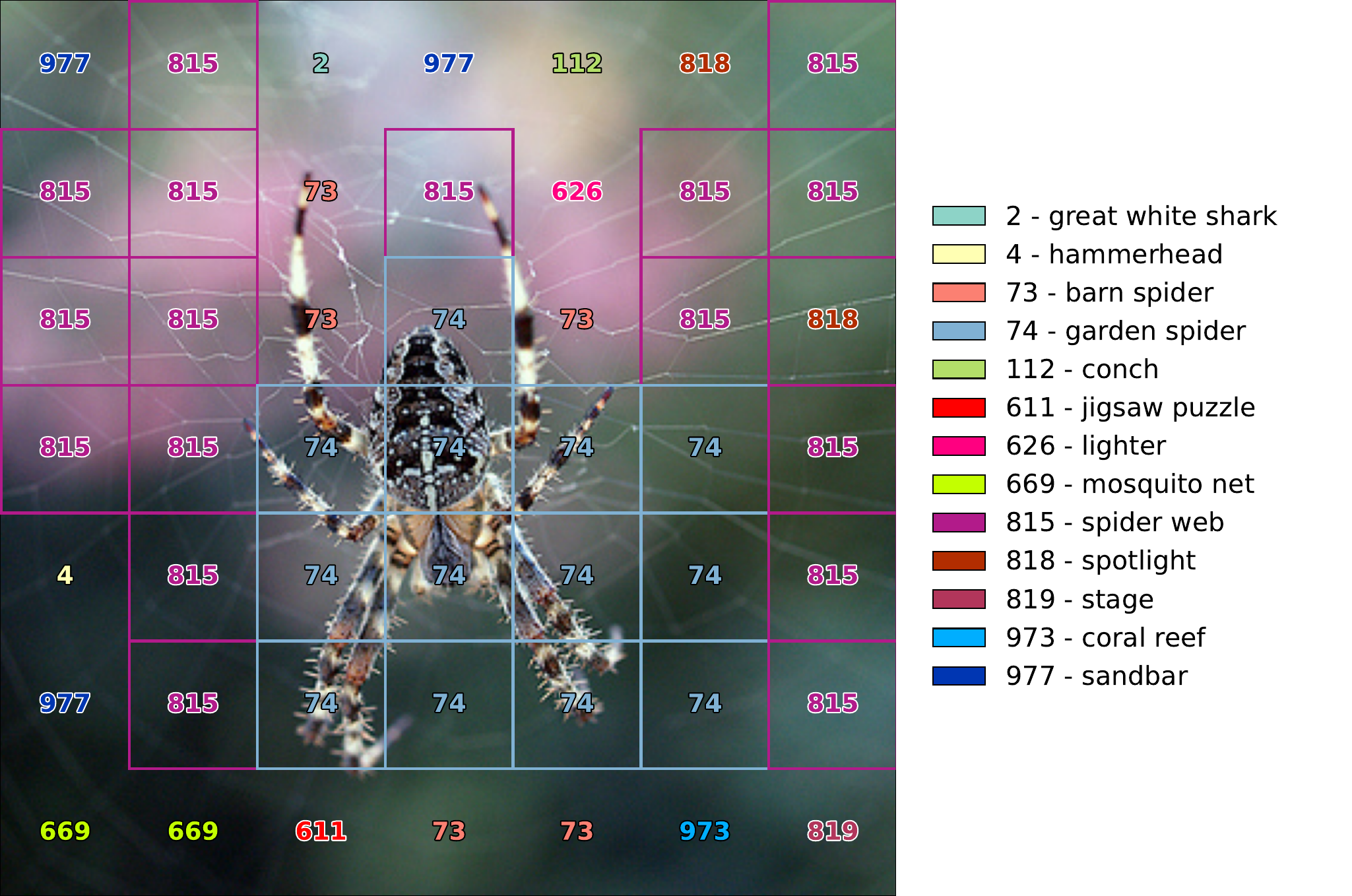}
    \caption{Original image}
    \label{fig:spider-original}
  \end{subfigure}
  \hfill
  \begin{subfigure}[b]{0.245\textwidth}
    \centering
    \includegraphics[width=\linewidth, height=4cm, keepaspectratio]{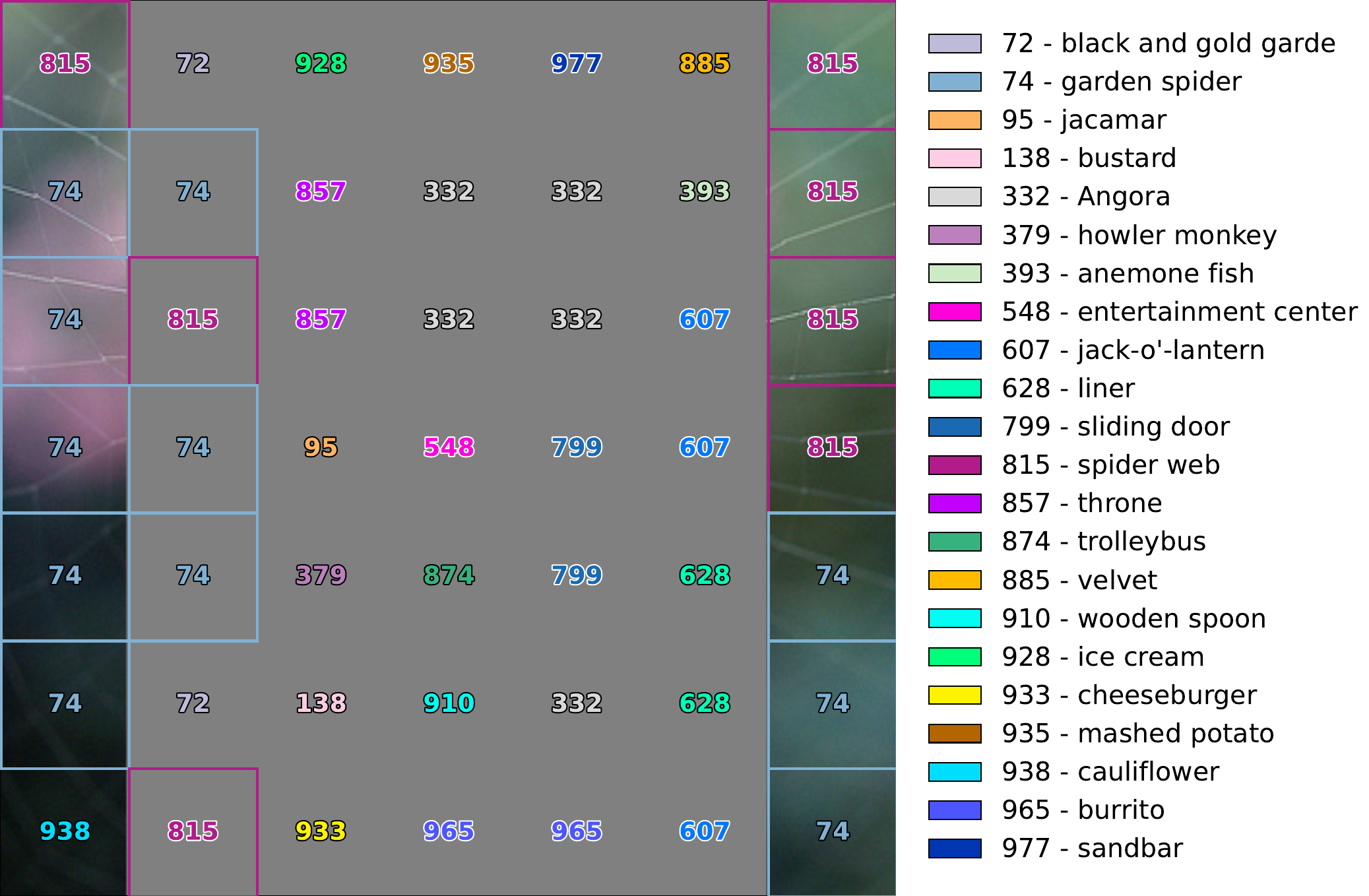}
    \caption{Occluded image}
    \label{fig:spider-occluded}
  \end{subfigure}
  \hfill
  \begin{subfigure}[b]{0.245\textwidth}
    \centering
    \includegraphics[width=\linewidth, height=4cm, keepaspectratio]{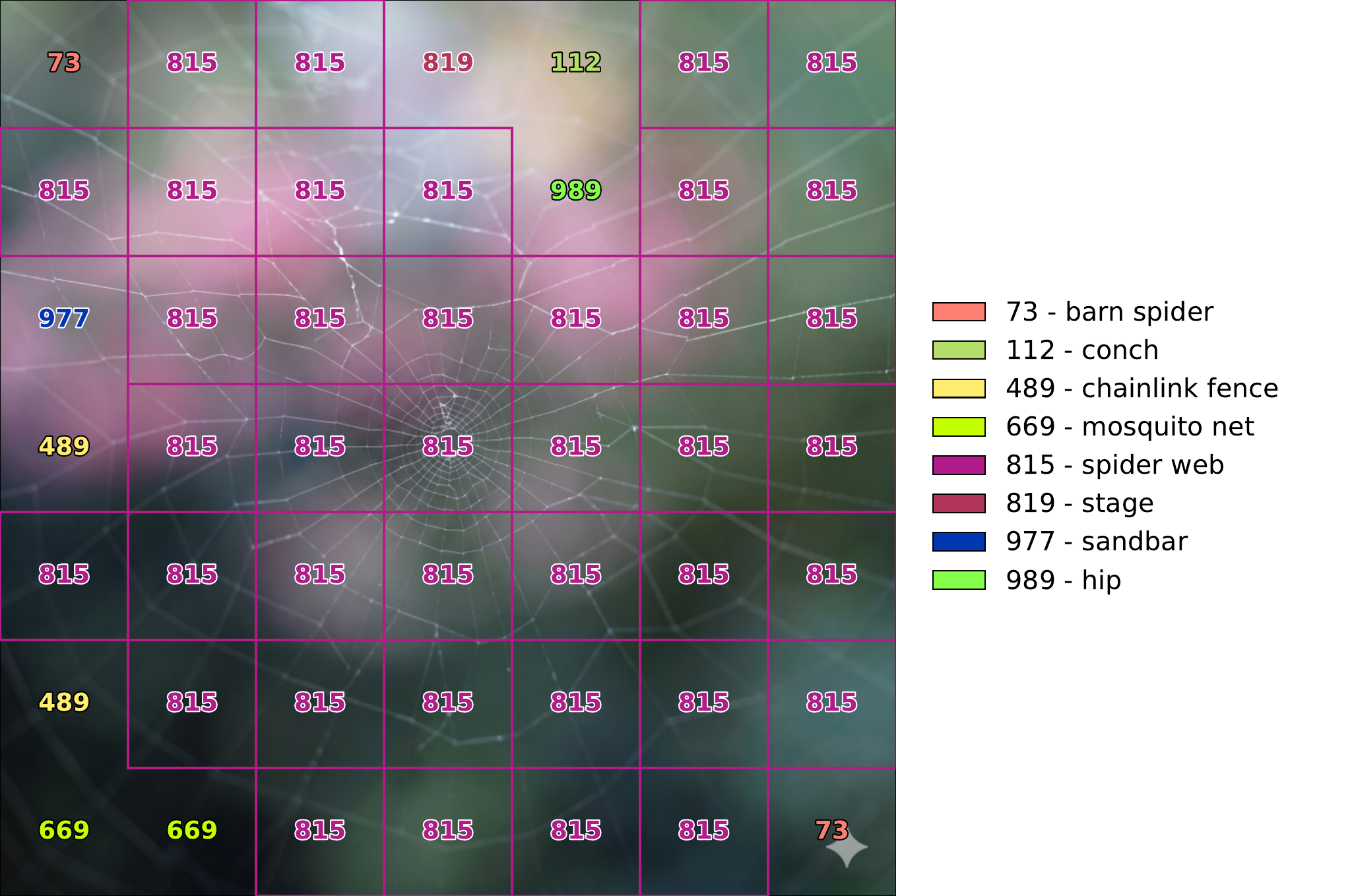}
    \caption{Inpainted image}
    \label{fig:spider-web}
  \end{subfigure}
  \begin{subfigure}[b]{0.245\textwidth}
    \centering
    \includegraphics[width=\linewidth, height=4cm, keepaspectratio]{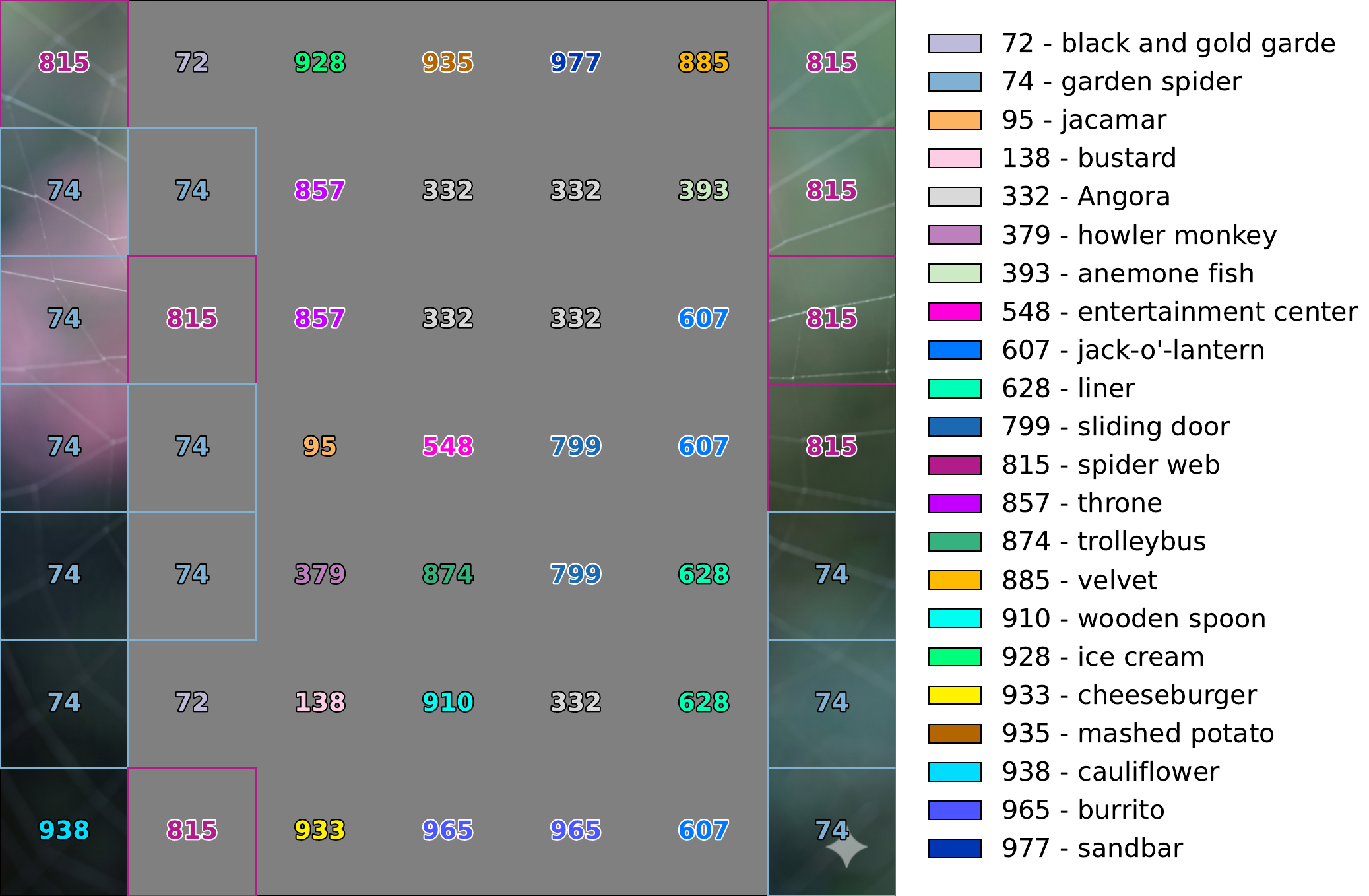}
    \caption{Occluded inpainted image}
    \label{fig:spider-web-occluded}
  \end{subfigure}
  \caption{Preliminary visual ablation of representation shifts in the \texttt{garden spider} class. (\subref{fig:spider-original}) Original image containing a spider. (\subref{fig:spider-occluded}) Foreground-occluded variant. (\subref{fig:spider-web}) Inpainted variant where the spider is removed using a generative model. (\subref{fig:spider-web-occluded}) Inpainted image with the same occlusion mask applied. On the images (\subref{fig:spider-original}, \subref{fig:spider-web}), background patches correctly predict \texttt{spider web}. However, introducing the occlusion mask (\subref{fig:spider-occluded}, \subref{fig:spider-web-occluded}) causes a subset of background patches to incorrectly switch to \texttt{garden spider}.}
  \label{fig:spider-occlusion}
\end{figure*}

The \texttt{garden spider} case illustrates a spatially resolved interaction between related class scores. The model assigns high local scores to web regions for \texttt{spider web} and to spider regions for \texttt{garden spider}. Under foreground occlusion, some web regions switch from \texttt{spider web} to \texttt{garden spider} (Fig.~\ref{fig:spider-occlusion}). The image-level prediction remains \texttt{garden spider} in both conditions, so this change is observable only at the spatial level. Whether this behavior reflects class-pair entanglement in the linear head, masking-induced distribution shift, or asymmetric feature dependencies between the two classes remains open. Additional details are provided in Appendix~\ref{app:context}.

These spatial effects are closely tied to the aggregation rule, and the MIL perspective reframes several image-classification limitations as aggregation failures. GAP should be treated as one specific aggregation assumption among others. Prior work shows that mean aggregation is not always optimal for weakly supervised learning, especially when image-level labels depend on sparse, uneven, or spatially competing evidence~\cite{carbonneau2018multiple,oquab2015localization,kraus2016classifying,durand2016weldon,durand2017wildcat,ilse2018attention}. Standard GAP classifiers may exhibit analogous limitations. At inference time, global averaging can dilute localized evidence that remains visible in the dense readout. During training, the same mean-pooled objective can distribute class supervision across foreground, part-level, and contextual features, encouraging spatial biases that are not apparent from image-level predictions alone. Alternative pooling rules such as max pooling, top-$k$ pooling, and attention-based MIL aggregation would change both the training dynamics and the interpretability of spatial evidence. Such rules should therefore be selected with the target failure mode in mind instead of being adopted as drop-in replacements.

Dense readouts also reveal architecture-dependent behavior. ViT-based models show high target-class activation in background cells on unmodified images and large drops after foreground occlusion, whereas CNNs show lower background activation and smaller drops. This pattern is consistent with object evidence being more spatially distributed in transformer-based models, whereas CNN evidence remains more tied to local receptive fields~\cite{raghu2021do}. It motivates further study of architecture-specific spatial dynamics and context dependence.

As a practical application, dense readouts may support adversarial attack debugging by isolating the local perturbations that drive an attack class (Fig.~\ref{fig:adversarial-attack} in Appendix~\ref{app:adversarial}). Combined with foreground occlusion, the dense readout also enables per-class measurements of context dependence and can help detect shortcut behavior when context alone triggers the target class in the absence of the object~\cite{geirhos2020shortcut,xiao2020noise}.

Nonlinear classification heads admit the same pointwise evaluation on spatial feature vectors, but the resulting spatial scores are no longer an exact decomposition of the aggregated image-level logits. A related observation holds for ViTs trained with \texttt{[CLS]} aggregation, where applying the classifier to patch-token representations can produce spatially localized class evidence despite the absence of GAP training. These results suggest that dense readouts may extend beyond the linear-GAP case, although their interpretation is less direct and requires separate analysis. Additional evidence is provided in Appendix~\ref{app:nonlinear-head}.

\subsection{Limitations.}

The spatial class-score maps are constrained by several limitations. First, the resolution is limited by the backbone's final feature output, which yields a relatively coarse grid for most standard networks and restricts precise object-boundary delineation. Attribution methods such as Layer-wise Relevance Propagation~\cite{bach2015pixel} can refine the spatial responses back to pixel space (Fig.~\ref{fig:lrp} in Appendix~\ref{app:lrp}). Second, standard pretrained classifiers lack an explicit background class, which forces the network to project semantically empty regions into the predefined target label space. Preliminary analysis suggests that confidence thresholding can filter some low-confidence artifacts, but it does not remove confident background activations (Appendix~\ref{app:uncertainty-thresholding}).

The spatial class scores should be interpreted as diagnostic evidence rather than calibrated local labels. Metrics such as \emph{FG Det.}\ require only one correct foreground cell, while \emph{Loc.}\ depends on coarse boxes or downsampled masks. Finally, treating spatial feature vectors as independent instances remains an approximation. Hierarchical receptive fields in CNNs aggregate information from overlapping neighborhoods (Fig.~\ref{fig:lrp} in Appendix~\ref{app:lrp}), and self-attention in ViTs redistributes information across tokens by construction~\cite{raghu2021do}. In ViT-based models, foreground occlusion reduces background activation by 52.4--53.4 points (Table~\ref{tab:bbox-eval}), suggesting that background tokens can reflect global object context rather than purely local features (Fig.~\ref{fig:warthog-occlusion} in the Appendix). Because occlusion also introduces artificial distribution shifts, context-dependence measurements should be interpreted cautiously.

\subsection{Future work.}

The background activations observed in our experiments highlight a limitation of standard classifiers, which lack a spatial background or reject option and therefore score every spatial feature vector against the foreground class set. Under the MIL reinterpretation, an image is a bag of instances, many of which often correspond to background. GAP aggregates these instances together with sparse discriminative evidence, allowing unsupported regions to influence the image-level prediction. This suggests a learned background-instance component that gives non-target instances a separate role in the aggregation process. The form of this component may depend on the architecture. In convolutional models, it could be implemented directly over the spatial instances exposed by the dense readout. In transformer-based models, self-attention makes the routing problem more flexible by allowing patch tokens to exchange foreground and background evidence before pooling. Prior work shows that additional tokens can absorb non-semantic information otherwise carried by patch tokens~\citep{darcet2024vision}. This mechanism could be extended toward explicit non-target or sink instances. The open problem is to learn such mechanisms from image-level supervision.

Dense spatial features also suggest extensions beyond supervised classification. For self-supervised models, spatial attribution can be defined by similarity to reference embeddings instead of classifier logits (Fig.~\ref{fig:similarity-connections} in Appendix~\ref{sec:appendix-applications}). This may provide a lightweight alternative to methods such as BiLRP~\cite{eberle2022building} when assessing whether a backbone encodes features relevant to a downstream task. In this setting, a small annotated reference set could be used to probe spatial correspondence without training a separate linear head.

Finally, the MIL perspective may inform dense self-supervised learning objectives. Instead of treating an image as a single training example, such objectives could model each image as a bag of spatial instances whose representations should remain consistent across augmented views. This formulation is consistent with dense SSL methods showing that local or region-level alignment can improve representation learning~\citep{wang2021densecl,xie2021propagate,xiao2021regions}. More broadly, it suggests a route toward simpler spatial SSL objectives that operate directly on native spatial feature maps and reduce reliance on dataset-level context or image-level shortcuts.

% In the unusual situation where you want a paper to appear in the
% references without citing it in the main text, use \nocite
% \nocite{langley00}

\bibliography{ref}
\bibliographystyle{icml2024}

%%%%%%%%%%%%%%%%%%%%%%%%%%%%%%%%%%%%%%%%%%%%%%%%%%%%%%%%%%%%%%%%%%%%%%%%%%%%%%%
%%%%%%%%%%%%%%%%%%%%%%%%%%%%%%%%%%%%%%%%%%%%%%%%%%%%%%%%%%%%%%%%%%%%%%%%%%%%%%%
% APPENDIX
%%%%%%%%%%%%%%%%%%%%%%%%%%%%%%%%%%%%%%%%%%%%%%%%%%%%%%%%%%%%%%%%%%%%%%%%%%%%%%%
%%%%%%%%%%%%%%%%%%%%%%%%%%%%%%%%%%%%%%%%%%%%%%%%%%%%%%%%%%%%%%%%%%%%%%%%%%%%%%%
\newpage
\appendix
\onecolumn

\section{Appendix}

\subsection{Natural Image Experiment Details}
\label{app:experimental-details}

All experiments use $224{\times}224$ inputs obtained by resizing the shorter side to 256 pixels, center-cropping to $224{\times}224$, and applying ImageNet mean and standard-deviation normalization. Pixel masks are transformed with the same resize and center crop using nearest-neighbor interpolation. For a model with grid size $G$, a grid cell is foreground if any foreground pixel falls inside the corresponding cell.

\begin{figure}[!t]
    \centering
    \includegraphics[width=0.6\textwidth]{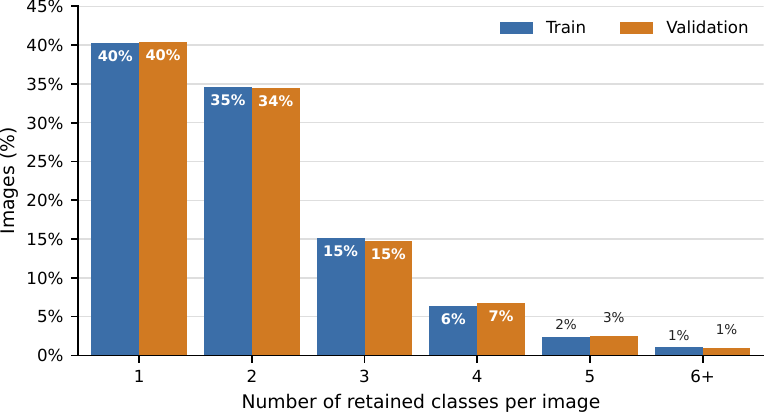}
    \caption{Distribution of the number of classes per image in the filtered COCO dataset.}
    \label{fig:coco_distribution}
\end{figure}

\paragraph{MS-COCO details.} We use the 2017 train and validation splits and filter instance annotations based on object visibility after preprocessing. Each segmentation mask is resized with nearest-neighbor interpolation to a short side of 256 pixels and center-cropped to \(224 \times 224\); annotations are retained only if the resulting mask occupies at least 1\% of the $224 \times 224$ crop. Images with at least one retained annotation are kept, and multi-label targets are constructed from the remaining object categories. This procedure preserves all 80 COCO classes and yields 112,343 training images and 4,737 validation images. The resulting dataset remains substantially multi-label: 67,121 training images (59.75\%) and 2,823 validation images (59.59\%) contain more than one retained class, with most images containing one to three classes, as shown in Fig.~\ref{fig:coco_distribution}. Training uses single-label cross-entropy, sampling one positive class uniformly from the set of classes present in each image. Validation uses a fixed deterministic positive label for single-label image accuracy.

\begin{table}[h]
\centering
\caption{Architectural properties of the encoders evaluated on MS-COCO at $224{\times}224$ input. \textbf{Grid} is the spatial resolution of the feature vectors, \textbf{Dim} is the spatial feature vector size before classification, and \textbf{Agg.} is the aggregation used by the encoder during pretraining to produce its global representation.}
\label{tab:coco-models}
\smallskip
\begin{tabular}{l l c c l}
\toprule
\textbf{Method} & \textbf{Backbone} & \textbf{Grid} & \textbf{Dim} & \textbf{Agg.} \\
\midrule
CLIP~\cite{radford2021learning}    & ResNet-50  & $7\times7$    & 2048 & AttnPool \\
CLIP~\cite{radford2021learning}    & ConvNeXt-B & $7\times7$    & 1024 & GAP \\
CLIP~\cite{radford2021learning}    & ViT-B/16   & $14\times14$  & 768  & \texttt{[CLS]} \\
DINOv1~\cite{caron2021emerging}    & ResNet-50  & $7\times7$    & 2048 & GAP \\
DINOv1~\cite{caron2021emerging}    & ViT-B/16   & $14\times14$  & 768  & \texttt{[CLS]} \\
DINOv2~\cite{oquab2023dinov2}      & ViT-B/14   & $16\times16$  & 768  & \texttt{[CLS]} \\
DINOv3~\cite{simeoni2025dinov3}    & ConvNeXt-B & $7\times7$    & 1024 & GAP \\
DINOv3~\cite{simeoni2025dinov3}    & ViT-B/16   & $14\times14$  & 768  & \texttt{[CLS]} \\
MAE~\cite{he2022masked}            & ViT-B/16   & $14\times14$  & 768  & \texttt{[CLS]} \\
BEiT-3~\cite{wang2023image}        & ViT-B/16   & $14\times14$  & 768  & \texttt{[CLS]} \\
\bottomrule
\end{tabular}%
\end{table}

\begin{table}[h]
\centering
\caption{Architectural properties for ImageNet bounding-box evaluation at $224{\times}224$ input. \textbf{Grid} is the spatial resolution of the feature vectors, \textbf{Dim} is the spatial feature vector size before classification, and \textbf{Agg.} denotes the image-level aggregation used during training.}
\label{tab:imagenet-models}
\smallskip
\begin{tabular}{l l c c l}
\toprule
\textbf{Model} & \textbf{Architecture} & \textbf{Grid} & \textbf{Dim} & \textbf{Agg.} \\
\midrule
EfficientNet-B0~\cite{tan2019efficientnet} & CNN (MBConv) & $7\times7$ & 1280 & GAP \\
ResNet-50~\cite{he2016deep} & CNN (residual) & $7\times7$ & 2048 & GAP  \\
ConvNeXt-B~\cite{liu2022convnet} & CNN & $7\times7$ & 1024 & GAP \\
Swin-B~\cite{liu2021swin} & hierarchical transformer & $7\times7$ & 1024 & GAP \\
MaxViT-B~\cite{tu2022maxvit} & hybrid conv-attention & $7\times7$ & 768 & GAP \\
\bottomrule
\end{tabular}%
\end{table}

\subsection{Synthetic Dataset Details}
\label{app:synmil-details}

\begin{figure}[!t]
  \centering
  \includegraphics[width=0.60\textwidth]{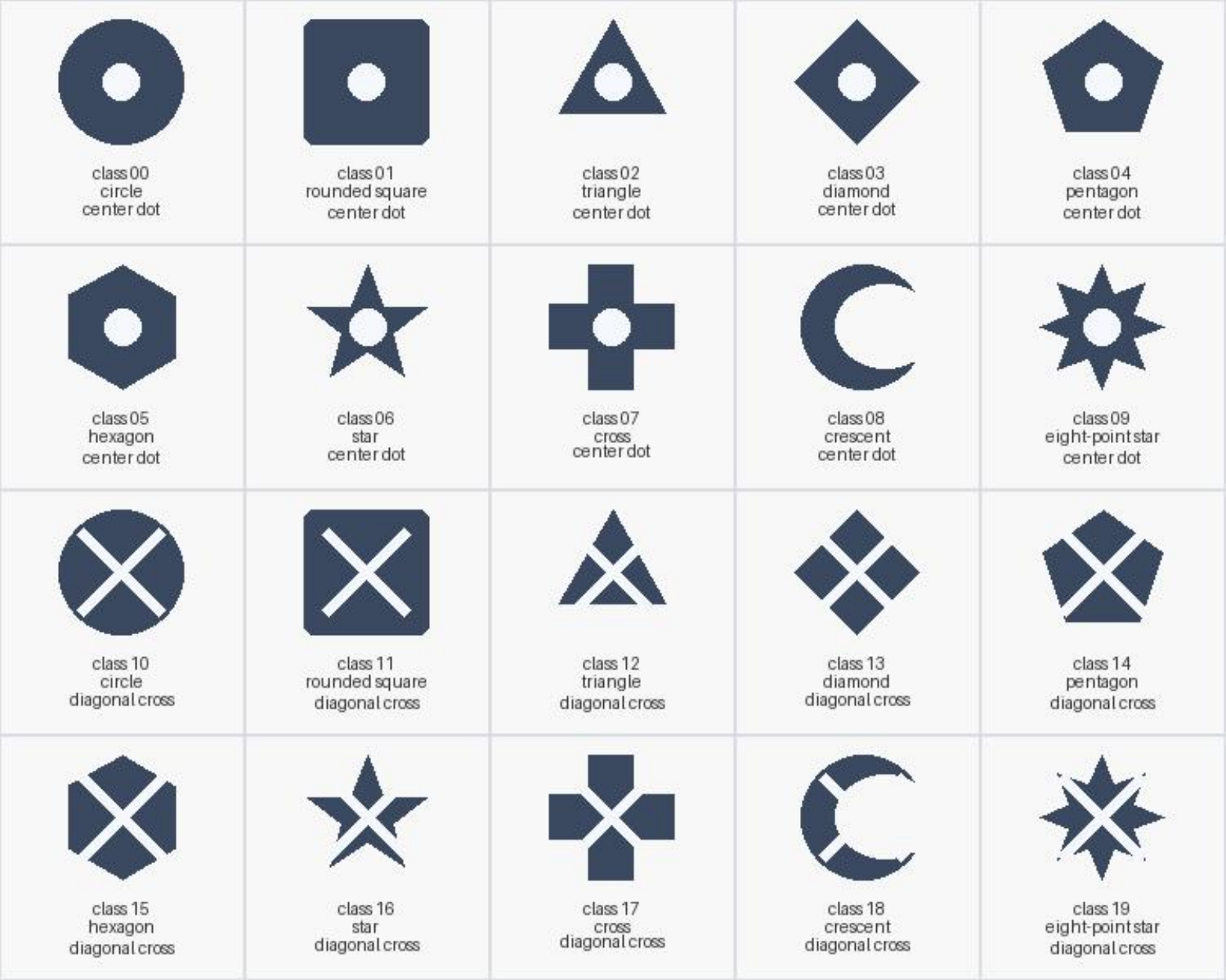}
  \caption{The 20 geometric classes of the synthetic dataset. Classes 0--9 share the center-dot pattern and classes 10--19 reuse the same shapes with a diagonal-cross pattern.}
  \label{fig:synthetic-classes}
\end{figure}

Images are $224{\times}224$ RGB scenes with 3--5 visible objects sampled without replacement from $K=20$ classes (Fig.~\ref{fig:synthetic-classes}). Class identity is encoded by a geometric symbol defined by the Cartesian product of ten base shapes and class-dependent internal patterns. For each image, the number of objects is sampled uniformly from $\{3,4,5\}$, and one visible class is then sampled uniformly and used as the sole image-level training target. Non-target objects are independently sampled co-occurring positives and are not predictive of the target beyond their presence in the same image. Object scale is sampled uniformly from 42 to 68 pixels, rotation uniformly from $[-180^\circ,180^\circ]$, and location uniformly subject to an overlap constraint, with randomized rendering order. Object color is sampled independently of class from a fixed palette with random channel jitter, and the background is sampled independently as a light RGB color. For each sample, the generator stores the RGB image, a semantic mask with label 0 denoting background and labels $1,\ldots,K$ denoting object classes, an instance mask with positive labels denoting per-image instances, and a JSON record containing the sampled target, the full visible class set, object bounding boxes, centers, areas, colors, rotations, and instance identifiers. Models are trained only with the sampled single image-level target, and the masks are used exclusively for evaluation.

The main synthetic run uses a 500k-image training set with 50k-image validation and test splits generated with seed 0. We train two randomly initialized GAP classifiers with mean pooling over spatial feature vectors. We use a standard ResNet-18 architecture. ViT-S/32 is implemented as \texttt{vit\_small\_patch32\_224} with no class token, average pooling, and no dropout or stochastic depth. For both ResNet-18 and ViT-S/32, the linear classifier is additionally applied before global pooling to obtain spatial class scores for validation metrics, while the training loss is applied only to the image-level output after pooling. Both models are trained with image-level cross-entropy for up to 100 epochs, using batch size 512 and deterministic data loading. ResNet-18 uses SGD with learning rate 0.15, momentum 0.9, weight decay $10^{-4}$, 5 warmup epochs, and cosine decay. ViT-S/32 uses AdamW with learning rate $1.5{\times}10^{-3}$, weight decay 0.05, 5 warmup epochs, and cosine decay to $10^{-5}$. Checkpoints are selected by validation performance, yielding epoch 14 for ResNet-18 and epoch 32 for ViT-S/32.

Test-set \emph{Top-1 Acc.} is measured against the sampled image-level target. For \emph{Loc.}, semantic masks are projected to the model grid using any-pixel occupancy. A foreground cell may contain more than one valid object class after projection, and a dense prediction is counted as correct if it matches any visible object class present in that cell. We omit foreground/background detection and occlusion metrics for this experiment because the multi-object setting makes their single-target interpretation less direct than in ImageNet and MS-COCO.

The lower ViT-S/32 localization value should be interpreted in the context of the lightweight training recipe used for this synthetic experiment. ViTs trained from random initialization are sensitive to data scale and optimization choices~\cite{dosovitskiy2020image}, and a full ViT-specific training sweep was outside the scope of this experiment. The validation logs are consistent with this limitation. ViT-S/32 reaches image-level accuracy near 26\%, but its localization metrics remain lower and decline under extended training, with \emph{Loc.}\ decreasing from 63\% at epoch 32 to 35.5\% at epoch 74. We therefore do not interpret the ResNet-18--ViT-S/32 difference as an architectural ranking.

\subsection{Uncertainty Thresholding Controls}
\label{app:uncertainty-thresholding}

\begin{figure}[th]
  \centering
  \includegraphics[width=\textwidth]{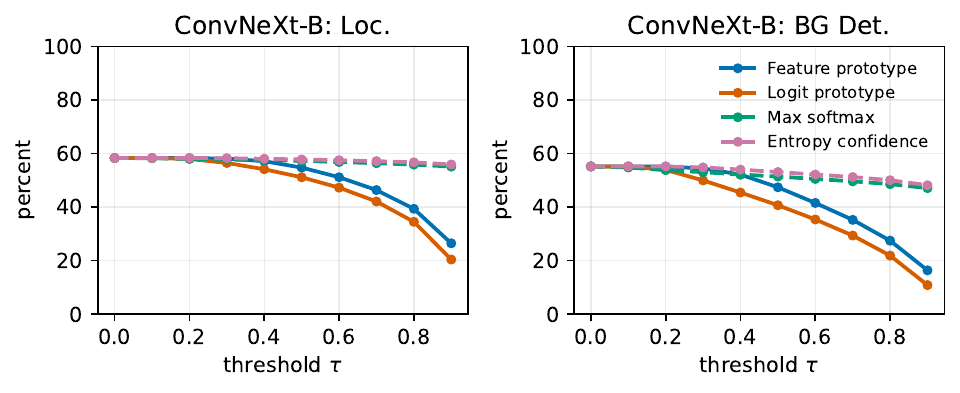}
  \caption{Effect of thresholding on ConvNeXt-B ImageNet spatial class-score maps. Curves show top-1 \emph{Loc.} and \emph{BG Act.} as the retention threshold $\tau$ increases for foreground-prototype and confidence-score methods.}
  \label{fig:uncertainty-thresholding-convnext}
\end{figure}

\begin{figure}[th]
  \centering
  \includegraphics[width=\textwidth]{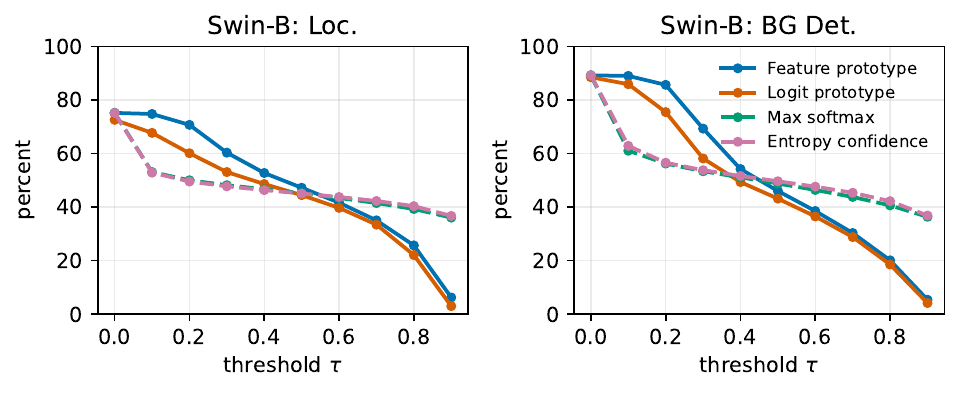}
  \caption{Effect of thresholding on Swin-B ImageNet spatial class-score maps. The same four thresholding methods are evaluated as in Figure~\ref{fig:uncertainty-thresholding-convnext}.}
  \label{fig:uncertainty-thresholding-swin}
\end{figure}

We additionally evaluate whether thresholding spatial predictions can reduce background activation without erasing foreground localization. For each grid cell, we compute a scalar score and retain the cell only when its score exceeds threshold $\tau$. We compare maximum softmax, normalized entropy confidence $1-H(p)/\log K$, and two foreground-prototype scores: cosine similarity to a feature-space prototype or logit-space prototype estimated from correctly predicted cells inside bounding boxes. Figure~\ref{fig:uncertainty-thresholding-convnext} and Figure~\ref{fig:uncertainty-thresholding-swin} report only \emph{Loc.} and background activation (\emph{BG Act.}), the two metrics needed to assess this tradeoff. We omit the energy-sigmoid score because it is flat across thresholds in these experiments.

The curves show that thresholding is not a free correction for spatial prediction noise. For ConvNeXt-B, \texttt{logit\_fg} at $\tau=0.4$ lowers \emph{BG Act.} from $55.2\%$ to $45.4\%$ while \emph{Loc.} drops from $58.3\%$ to $54.1\%$. For Swin-B, preserving \emph{Loc.} leaves most background activation intact: for example, \texttt{feat\_fg} at $\tau=0.2$ lowers \emph{BG Act.} from $89.2\%$ to $85.7\%$ while \emph{Loc.} drops from $75.2\%$ to $70.7\%$, and stronger thresholds quickly remove foreground evidence. Therefore, while thresholding can reduce background activation, it does so at the cost of also reducing foreground localization, and it does not fully close the gap between these metrics. This suggests that improving spatial class scores may require architectural changes to reduce noise or better uncertainty estimation, rather than post-hoc thresholding alone.

\subsection{Nonlinear Heads and \texttt{[CLS]} Aggregation}
\label{app:nonlinear-head}

Table~\ref{tab:nonlinear-heads} evaluates whether pointwise spatial readout remains informative beyond the exact linear-GAP decomposition. We train four ImageNet-1K classifiers from random initialization: ResNet-18 and ViT-S/32 with mean pooling over the final spatial or token grid, each paired with either a linear fully connected (FC) head or a three-layer MLP head. The ViT-S/32 model is implemented as \texttt{vit\_small\_patch32\_224} without a class token and with average pooling over its $7{\times}7$ patch-token grid. We additionally evaluate a pretrained \texttt{timm} ViT-B/16 classifier trained with \texttt{[CLS]}. For the pretrained ViT-B/16, the image-level prediction is produced from the class token, while the spatial readout is obtained by applying the classifier to final patch-token representations. All models are evaluated using the same bounding-box protocol as Table~\ref{tab:bbox-eval}.

For the MLP condition, only the final classifier is replaced. Given pre-classifier feature dimension $d$, the head consists of \(\operatorname{Linear}(d,d)\), batch normalization, ReLU, \(\operatorname{Linear}(d,d)\), batch normalization, ReLU, and \(\operatorname{Linear}(d,1000)\). Thus, $d=512$ for ResNet-18 and $d=384$ for ViT-S/32. The MLP head contains no dropout. At evaluation, the same head is applied pointwise to each final convolutional cell or patch-token representation to obtain spatial class scores.

The ResNet-18 MLP preserves image-level accuracy but substantially changes the spatial readout. Top-1 accuracy increases slightly from 70.5\% to 71.4\%, and foreground detection remains high at 83.4\%. However, foreground-cell localization drops from 52.8\% to 16.3\%, indicating that correct target predictions persist, but only on a smaller subset of foreground cells. One possible explanation is that the MLP relies on feature combinations available only after pooling. A pooled representation can combine multiple object cues across spatial locations, whereas a single foreground feature vector may contain only partial evidence.

The ViT-S/32 MLP shows a different pattern. Localization changes only mildly relative to the linear head, from 49.7\% to 47.4\%, while foreground detection and ReaL recovery remain essentially unchanged. This stability likely reflects self-attention, which contextualizes final-layer patch tokens before pooling. As a result, pointwise classification can remain informative under both linear and nonlinear heads. This interpretation is also compatible with the high background activation observed in ViT models, where background tokens can carry target-class evidence propagated from foreground tokens.

The pretrained \texttt{[CLS]} ViT-B/16 also yields strong spatial readouts from patch tokens, with 90.8\% foreground detection and 72.7\% foreground-cell localization. In \texttt{[CLS]}-based ViTs, the class token is updated through repeated attention-based aggregation over patch tokens, and the final classifier is trained on the resulting image-level representation~\cite{dosovitskiy2020image,raghu2021do}. Prior class-attention and weakly supervised localization methods exploit this aggregation pathway to recover spatial evidence~\cite{touvron2021going,xu2022multi}. Our empirical results do not establish a formal patch-level classifier, but they show that class-discriminative evidence remains accessible in the patch-token stream that contributes to the image-level prediction.

\begin{table}[th]
\centering
\scriptsize
\caption{ImageNet bounding-box evaluation for nonlinear-head and \texttt{[CLS]} controls. \emph{FC} denotes a linear classifier head, \emph{MLP} denotes a three-layer head, and \texttt{[CLS]} denotes class-token aggregation. Values are top-1 percentages.}
\label{tab:nonlinear-heads}
\smallskip
\begin{tabular*}{\textwidth}{@{\extracolsep{\fill}} l c c c c c c @{}}
\toprule
Model & \shortstack{Top-1\\Acc.} & \shortstack{FG\\Det.} & Loc. & \shortstack{BG\\Act.} & \shortstack{BG Act.\\(w/o FG)} & \shortstack{ReaL\\Rec.} \\
\midrule
ResNet-18 FC  & 70.5 & 87.3 & 52.8 & 60.8 & 14.9 & 85.1 \\
ResNet-18 MLP & 71.4 & 83.4 & 16.3 & 21.1 & 15.1 & 82.4 \\
ViT-S/32 FC   & 70.3 & 89.6 & 49.7 & 71.9 & 22.1 & 87.9 \\
ViT-S/32 MLP  & 70.1 & 89.2 & 47.4 & 70.2 & 21.9 & 87.9 \\
\midrule
ViT-B/16 \texttt{[CLS]} & 75.7 & 90.8 & 72.7 & 86.2 & 24.4 & 86.6 \\
\bottomrule
\end{tabular*}
\end{table}

\subsection{Class Activation Mapping for ViTs}\label{app:vit-cam}

Although class activation mapping (CAM) was originally developed for CNNs, the same formulation extends directly to GAP-based ViTs, where the patch tokens preserve spatial correspondence to image regions. For a ViT with $N$ patch tokens, each token is embedded as a $C$-dimensional feature vector. The final image-level embedding is obtained by averaging these token embeddings across the spatial dimension, yielding a single $C$-dimensional vector that is fed into a linear classification head. This formulation allows each token's feature vector to be weighted by the corresponding classifier weights, producing a class-specific heatmap over the spatial token grid, analogous to CAM in CNNs. This approach enables visualization of the spatial distribution of class evidence in ViTs without requiring any architectural modifications or additional training, providing insight into how ViTs localize semantic information across the image.

% figure with 4 subfigs in a row: banana, egyptian cat, indigo bunting, strawberry
\begin{figure}[th]
  \centering
  \begin{subfigure}[t]{0.175\textwidth}
    \centering
    \includegraphics[width=\linewidth]{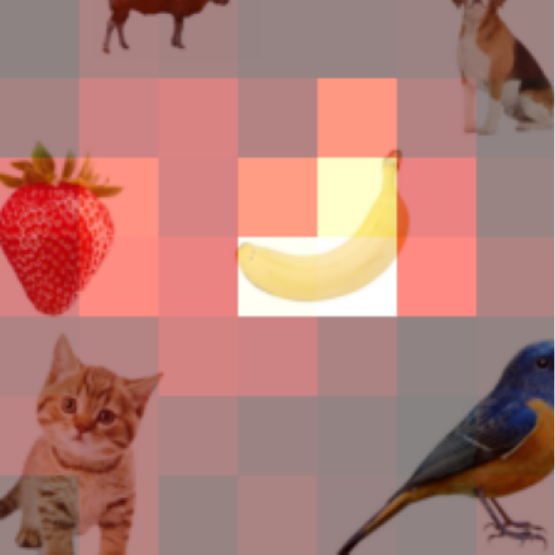}
    \caption{\texttt{banana}}
  \end{subfigure}
  \hfill
  \begin{subfigure}[t]{0.175\textwidth}
    \centering
    \includegraphics[width=\linewidth]{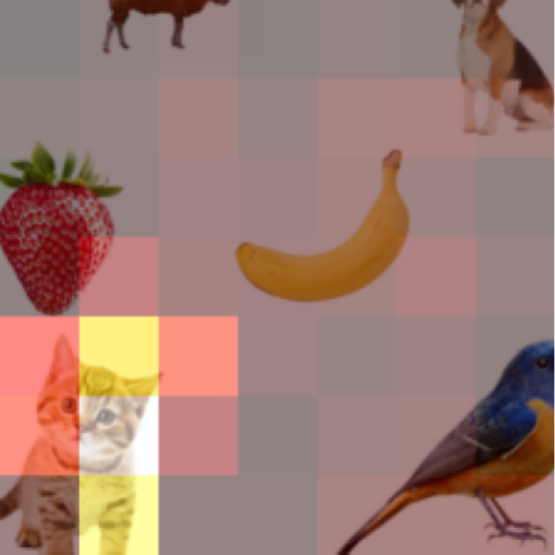}
    \caption{\texttt{Egyptian cat}}
  \end{subfigure}
  \hfill
  \begin{subfigure}[t]{0.175\textwidth}
    \centering
    \includegraphics[width=\linewidth]{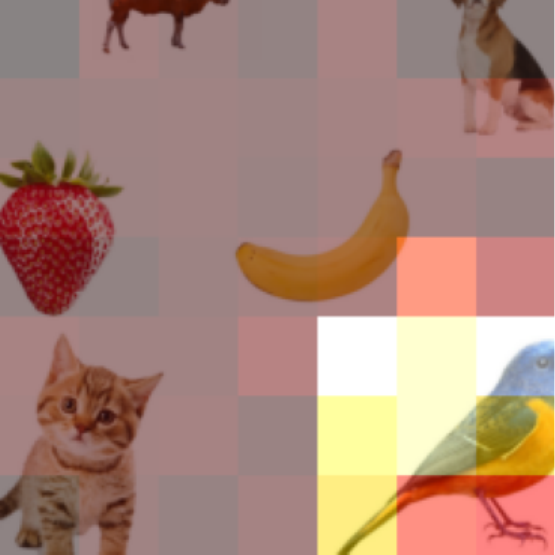}
    \caption{\texttt{bunting}}
  \end{subfigure}
  \hfill
  \begin{subfigure}[t]{0.175\textwidth}
    \centering
    \includegraphics[width=\linewidth]{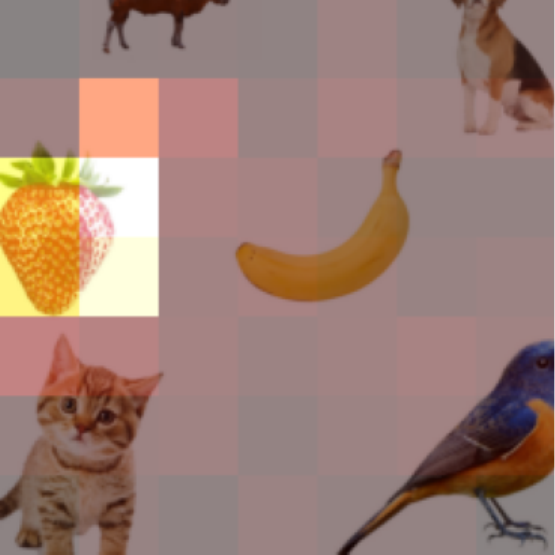}
    \caption{\texttt{strawberry}}
  \end{subfigure}
  \hfill
  \begin{subfigure}[t]{0.265\textwidth}
    \centering
    \includegraphics[width=\linewidth]{figures/SwinBaseIN1K_misc}
    \caption{Spatial class scores}
  \end{subfigure}
  \caption{Class activation maps for a pretrained Swin model.}
  \label{fig:vit-cam}

\end{figure}

\begin{table}[t]
\centering
\caption{Selected ImageNet classes whose background activation rate remains near 100\% even after foreground occlusion, across all architectures. These are classes whose context is highly diagnostic of the label. All values are percentages.}
\label{tab:vit-bg-classes}
\smallskip
\begin{tabular}{l l c c c c}
\toprule
\textbf{Model} & \textbf{Class} & \textbf{Loc.} & \textbf{FG Det.} & \textbf{BG Act.} & \textbf{BG Act.\ (w/o FG)} \\
\midrule
MaxViT       & web site               & 86.8 & 100.0 & 100.0 & 100.0 \\
MaxViT       & yellow lady's slipper  & 95.3 & 100.0 & 100.0 & 100.0 \\
MaxViT       & basketball             & 95.5 & 100.0 & 100.0 & \ 98.0 \\
MaxViT       & bearskin               & 95.9 & 100.0 & 100.0 & \ 98.0 \\
Swin         & web site               & 83.8 & 100.0 & 100.0 & 100.0 \\
Swin         & basketball             & 96.5 & 100.0 & 100.0 & \ 96.0 \\
Swin         & bearskin               & 96.8 & 100.0 & 100.0 & \ 96.0 \\
ViT          & basketball             & 92.5 & \ 92.3 & 100.0 & \ 96.0 \\
ViT          & volleyball             & 89.8 & 100.0 & 100.0 & \ 95.9 \\
ConvNeXt     & web site               & 69.4 & 100.0 & 100.0 & 100.0 \\
ConvNeXt     & yellow lady's slipper  & 83.4 & 100.0 & \ 90.2 & 100.0 \\
EfficientNet & web site               & 70.8 & 100.0 & 100.0 & 100.0 \\
ResNet-50    & web site               & 51.3 & 100.0 & 100.0 & 100.0 \\
\bottomrule
\end{tabular}%
\end{table}

\subsection{Class-Level Diagnostics of Context Dependence}
\label{app:context}

\begin{table}[!t]
\centering
\caption{Top 20 ResNet-50 ImageNet classes ranked by $\Delta = \text{BG Act.\ (w/o FG)} - \text{BG Act.}$ A positive $\Delta$ indicates that occluding the foreground \emph{increases} the rate of target-class predictions in the background. All values are percentages.}
\label{tab:cnn-bg-delta}
\smallskip
\begin{tabular}{r l c c c c c}
\toprule
\textbf{ID} & \textbf{Class} & \textbf{Loc.} & \textbf{FG Det.} & \textbf{BG Act.} & \textbf{BG Act.\ (w/o FG)} & \textbf{$\Delta$} \\
\midrule
\ 95 & jacamar                  & 42.1 & \ 98.0 & 33.3 & 71.8 & +38.5 \\
\ 74 & garden spider            & 27.6 & \ 82.0 & 33.3 & 66.7 & +33.3 \\
115  & sea slug                 & 48.5 & \ 98.0 & 57.6 & 87.9 & +30.3 \\
995  & earthstar                & 47.6 & 100.0 & 13.2 & 42.1 & +28.9 \\
139  & ruddy turnstone          & 48.7 & 100.0 & 22.7 & 50.0 & +27.3 \\
320  & damselfly                & 38.8 & 100.0 & 29.7 & 56.8 & +27.0 \\
\ 25 & European fire salamander & 45.7 & \ 98.0 & \ 6.7 & 33.3 & +26.7 \\
102  & echidna                  & 51.4 & 100.0 & 37.1 & 60.0 & +22.9 \\
\ \ 0 & tench                   & 52.3 & \ 94.0 & 32.7 & 55.1 & +22.4 \\
138  & bustard                  & 40.7 & 100.0 & 36.1 & 58.3 & +22.2 \\
\ 69 & trilobite                & 50.6 & \ 98.0 & 53.3 & 75.6 & +22.2 \\
\ 41 & whiptail                 & 34.9 & \ 98.0 & 17.9 & 38.5 & +20.5 \\
324  & cabbage butterfly        & 43.4 & \ 98.0 & 40.9 & 61.4 & +20.5 \\
809  & soup bowl                & 23.0 & \ 96.0 & \ 0.0 & 20.0 & +20.0 \\
\ 92 & bee eater                & 42.5 & \ 98.0 & 39.1 & 58.7 & +19.6 \\
365  & orangutan                & 33.7 & \ 94.0 & 14.3 & 33.3 & +19.0 \\
381  & spider monkey            & 16.6 & \ 80.0 & \ 3.7 & 22.2 & +18.5 \\
\ \ 3 & tiger shark             & 37.2 & \ 98.0 & 31.6 & 50.0 & +18.4 \\
991  & coral fungus             & 49.7 & \ 98.0 & 12.1 & 30.3 & +18.2 \\
984  & rapeseed                 & 70.0 & 100.0 & 53.3 & 71.1 & +17.8 \\
\bottomrule
\end{tabular}%
\end{table}

Evaluating the dense logit map under targeted occlusion provides a localized measurement of context dependence. This diagnostic can help isolate specific environmental features driving the classification and may detect potential shortcut learning when the background strictly determines the prediction~\cite{geirhos2020shortcut,xiao2020noise}.

Table~\ref{tab:vit-bg-classes} reports classes exhibiting near-perfect background activation across multiple architectures, largely independent of object presence. For these classes, the surrounding context appears to serve as a highly diagnostic proxy for the label~\cite{xiao2020noise}. Visual inspection suggests that \texttt{basketball} predictions in the background frequently coincide with correlated features such as players and the court. Similarly, background activations for the \texttt{bearskin} class seem to be triggered by the presence of military uniforms. For such classes, the dense readout indicates that the primary object may often be redundant for the final image-level prediction.

Table~\ref{tab:cnn-bg-delta} details a distinct phenomenon observed predominantly in CNNs: occluding the foreground paradoxically increases the rate of target-class predictions in the background. We quantify this representation shift via the metric $\Delta$, tracking the absolute increase in background activation upon foreground removal. The most affected classes are frequently wildlife tied to specific habitats (e.g., \texttt{jacamar} +38.5, \texttt{European fire salamander} +26.7) or artifacts defined by their context (e.g., \texttt{soup bowl} +20.0). For such categories, associated environmental features (e.g., specific foliage or accompanying tableware) may act as latent proxies that are partially suppressed when the dominant primary object is visible, but emerge when it is occluded.

To briefly explore this phenomenon, we conduct a preliminary visual ablation on the \texttt{garden spider} class (Fig.~\ref{fig:spider-occlusion}) using a ResNet-50 model. On the unmodified natural image, the model predicts \texttt{spider web} on background patches and \texttt{garden spider} directly on the insect. Upon occluding the foreground, the maximum logit on several adjacent web patches flips to \texttt{garden spider}.

To test whether the model simply uses the natural web as a proxy for the spider when the foreground is missing, we evaluate an inpainted variant where the spider is removed using the Nano Banana 2 generative model~\cite{raisinghani2026nano}. On this spider-free image, the patches are correctly classified as \texttt{spider web}. However, applying the same occlusion mask to the inpainted image once again causes the background patches to flip to \texttt{garden spider}. Changing the solid mask color yields identical representation shifts.

Although these results were collected for a single sample, they suggest that the shift in this instance is not driven solely by the natural background context. Instead, the behavior appears tied to the introduction of the artificial occlusion mask itself. Given that CNNs inherently rely on overlapping receptive fields, spatial vectors in the adjacent background inevitably encode the unnatural boundary of the mask. However, why this specific perturbation systematically triggers the occluded target class remains unclear. Whether this behavior stems from an out-of-distribution artifact interacting with local features, complex logit entanglement in the linear head, or distribution shifts within normalization layers remains an open question for future work.

\begin{figure}[!b]
  \centering
  \begin{subfigure}[b]{0.61\textwidth}
    \centering
    \includegraphics[width=\linewidth, keepaspectratio]{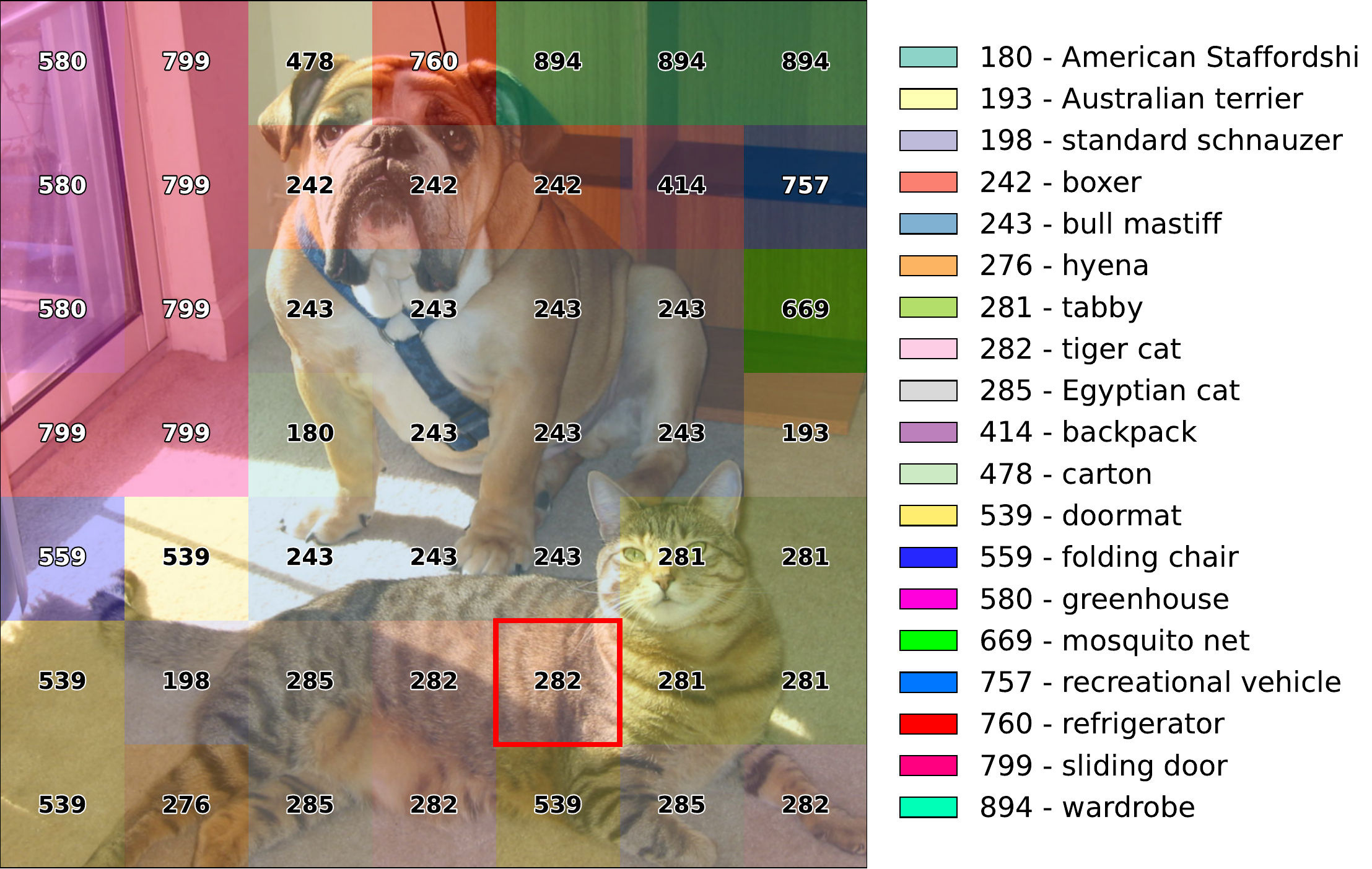}
    \caption{Spatial class scores}
  \end{subfigure}
  \hfill
  \begin{subfigure}[b]{0.385\textwidth}
    \centering
    \includegraphics[width=\linewidth, keepaspectratio]{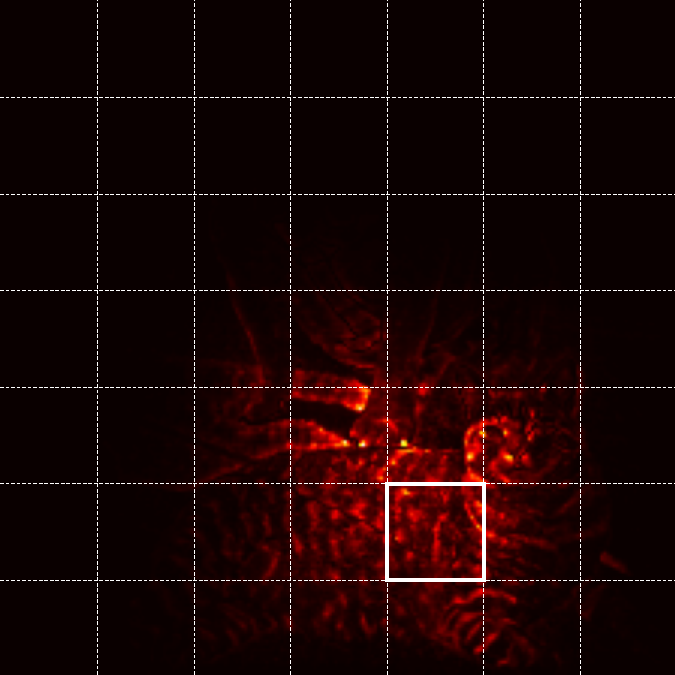}
    \caption{LRP saliency map}
  \end{subfigure}
  \caption{LRP saliency map for a target spatial class score highlighted in red (left) and white (right). The saliency map demonstrates receptive field overlap.}
  \label{fig:lrp}
\end{figure}

\subsection{Layer-wise Relevance Propagation for Dense Readout}
\label{app:lrp}

\begin{figure}[!b]
    \centering
    \includegraphics[width=\textwidth]{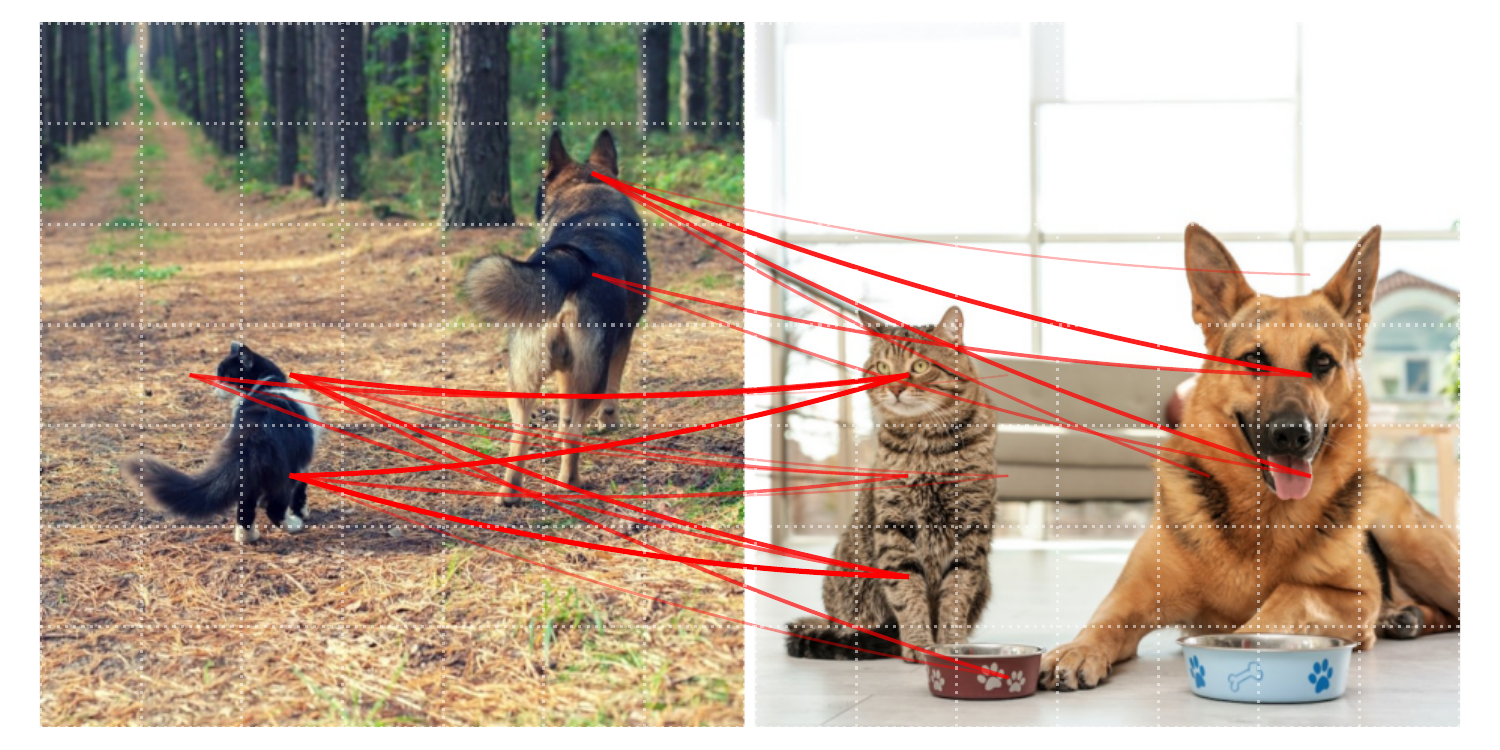}
    \caption{Patch-level spatial correspondence mapping between two images. Lines connect the top-20 most similar pairs of pre-aggregation spatial feature vectors, evaluated via cosine similarity in the latent space. This provides explicit similarity attribution by directly linking shared visual concepts across scenes.}
    \label{fig:similarity-connections}
\end{figure}

Layer-wise relevance propagation (LRP) is a post-hoc attribution method that decomposes a model's prediction back to the input space, assigning relevance scores to each input feature based on its contribution to the output. In the context of dense readout, LRP can be applied to the spatial class scores before global aggregation, allowing us to visualize which regions of the input image contribute most strongly to specific class predictions at the spatial level, as shown in Figure~\ref{fig:lrp}.

\subsection{Extended Applications of Pre-Aggregation Features}
\label{sec:appendix-applications}

\begin{figure}[!t]
  \centering
  \begin{subfigure}[b]{0.32\textwidth}
    \centering
    \includegraphics[width=\linewidth, keepaspectratio]{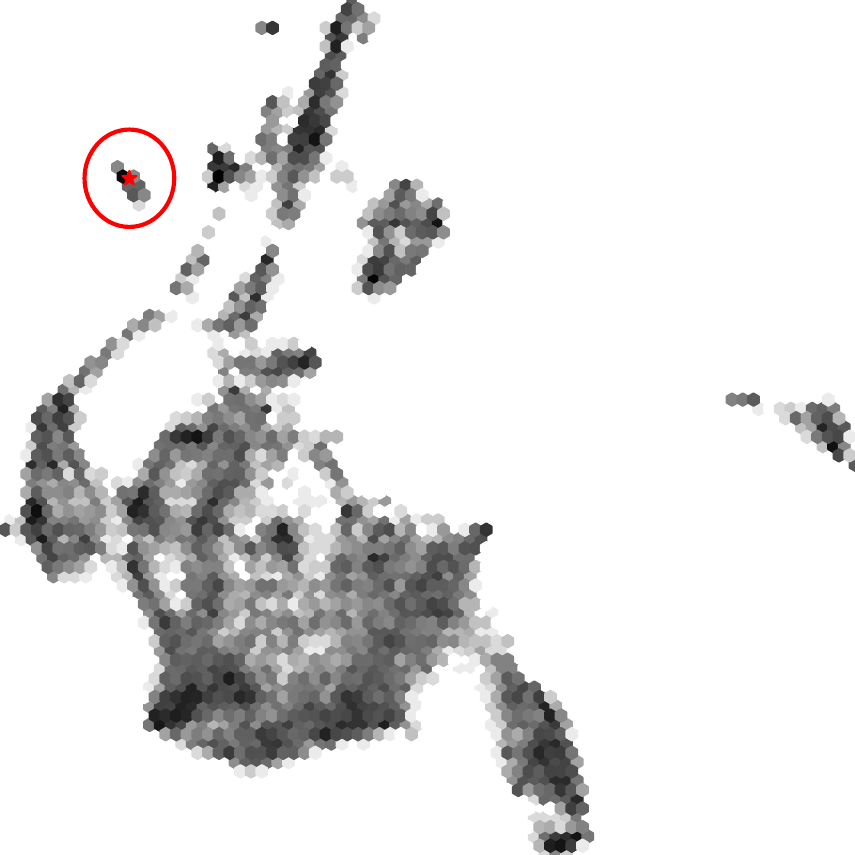}
    \caption{UMAP visualization}
    \label{fig:umap}
  \end{subfigure}
  \hfill
  \begin{subfigure}[b]{0.32\textwidth}
    \centering
    \includegraphics[width=\linewidth, keepaspectratio]{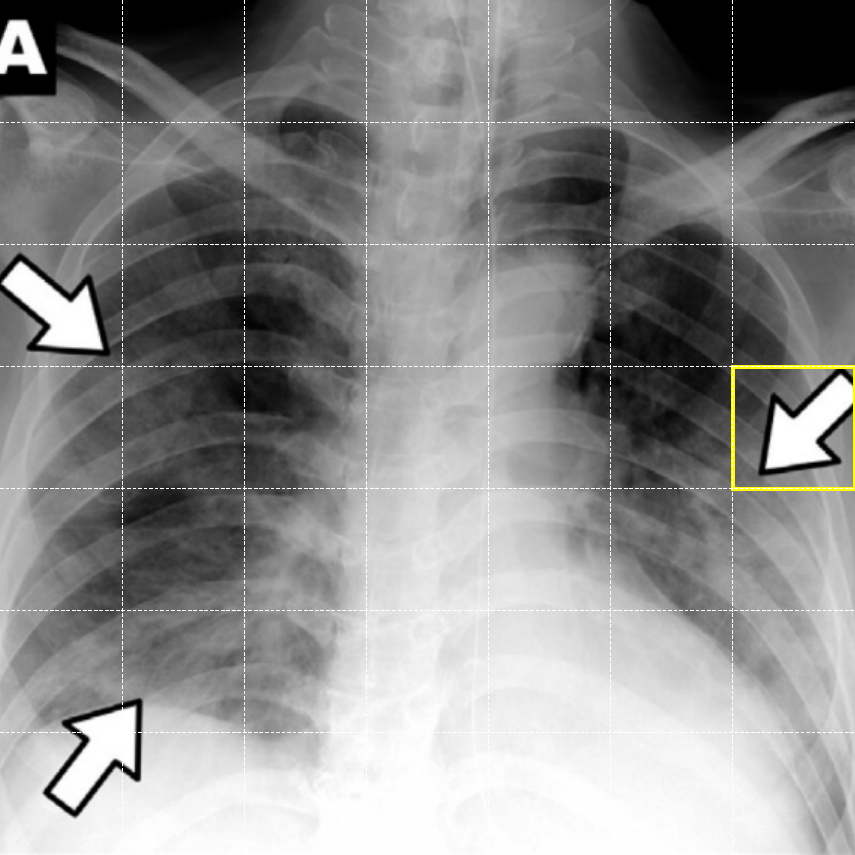}
    \caption{X-ray image}
    \label{fig:covid-slide}
  \end{subfigure}
  \hfill
  \begin{subfigure}[b]{0.32\textwidth}
    \centering
    \includegraphics[width=\linewidth, keepaspectratio]{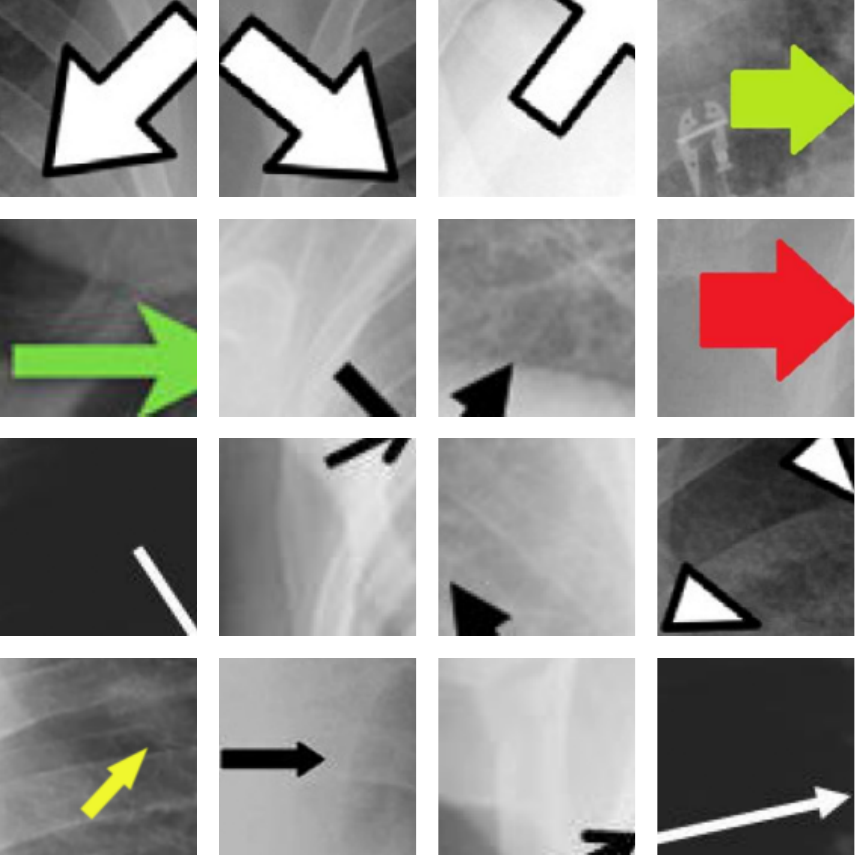}
    \caption{Artifact cluster}
    \label{fig:artefact-cluster}
  \end{subfigure}
  \caption{Spatial feature-vector analysis on the COVID-19 chest X-ray dataset~\cite{cohen2020covid}. (\subref{fig:umap}) UMAP projection of the spatial feature vectors extracted via the PubMedCLIP encoder prior to global aggregation. A distinct semantic cluster of clinical artifacts is highlighted (red circle), with a specific spatial location marked (red star). (\subref{fig:covid-slide}) Source radiograph corresponding to the marked location, mapping the spatial activation back to an annotation arrow. (\subref{fig:artefact-cluster}) Semantically similar patches retrieved from the highlighted artifact cluster across the dataset. A live interactive visualization is available at \url{https://karay.me/examples/umap_covid.html}.}
  \label{fig:patch-analysis}
\end{figure}

The preservation of spatial feature maps naturally extends beyond dense classification to fine-grained dataset analysis and similarity attribution. In this section, we provide two qualitative demonstrations of latent-space analysis using pre-aggregation spatial feature vectors.

Pre-aggregation spatial features naturally extend to similarity attribution between image pairs. Standard image similarity metrics evaluate the distance between globally averaged embeddings. This aggregation conceals the specific local correspondences driving the global similarity score. Bypassing global aggregation enables direct patch-level correspondence mapping.

We achieve this by computing the pairwise cosine similarity between all pre-aggregation spatial feature vectors from two given images. Connecting the highest-scoring pairs yields an explicit visual attribution map that links shared semantic concepts across the two scenes (Figure~\ref{fig:similarity-connections}). Because this approach leverages the inherently decomposable structure of the pre-aggregation grid, it operates strictly via a standard forward pass. Consequently, it serves as a direct and computationally efficient alternative to complex attribution frameworks such as Bilayer Layer-wise Relevance Propagation (BiLRP) for explaining pairwise similarity judgments.

Pre-aggregation spatial features also support fine-grained dataset analysis and diagnostics. Standard analytical approaches evaluate globally aggregated embeddings, which inevitably conflate multiple distinct objects into a single representation. By indexing the spatial feature vectors prior to GAP, we enable localized latent-space analysis without requiring bounding box annotations or a trained classification head.

We demonstrate this using the COVID-19 chest X-ray dataset~\cite{cohen2020covid}. Medical radiographs represent highly composite scenes where localized pathologies, medical hardware, and clinical annotations coexist. We extract pre-aggregation spatial activations using the frozen ResNet-50 encoder of PubMedCLIP~\cite{eslami2023pubmedclip}. Figure~\ref{fig:patch-analysis}\subref{fig:umap} illustrates a UMAP projection~\cite{mcinnes2018umap} of these spatial features across the dataset. Without global spatial aggregation, the latent representations naturally cluster by local semantic similarity rather than being diluted into a single averaged embedding.

This localized representation allows us to isolate highly specific dataset artifacts directly from the embedding space. For example, clinical artifacts such as arrows form a distinct, isolated cluster. Selecting a spatial vector from this cluster enables precise mapping back to the corresponding region in the source radiograph (Figure~\ref{fig:patch-analysis}\subref{fig:covid-slide}). Furthermore, querying nearest neighbors in this localized feature space isolates semantically homologous patches across the entire dataset (Figure~\ref{fig:patch-analysis}\subref{fig:artefact-cluster}). This provides a diagnostic tool for uncovering dataset biases that are otherwise obscured by global average pooling~\cite{torralba2011unbiased}.

\subsection{Spatial Decomposition of Adversarial Perturbations}
\label{app:adversarial}

Standard evaluations of adversarial robustness analyze perturbations strictly through the globally aggregated prediction. This image-level perspective obscures the underlying spatial dynamics of the attack. By retaining the pre-aggregation spatial feature vectors, we can decompose the adversarial optimization process and observe exactly how the attack corrupts the image at the patch level.

\begin{figure}[th]
  \centering
  \begin{subfigure}[b]{0.30\textwidth}
    \centering
    \includegraphics[width=\linewidth, height=4cm, keepaspectratio]{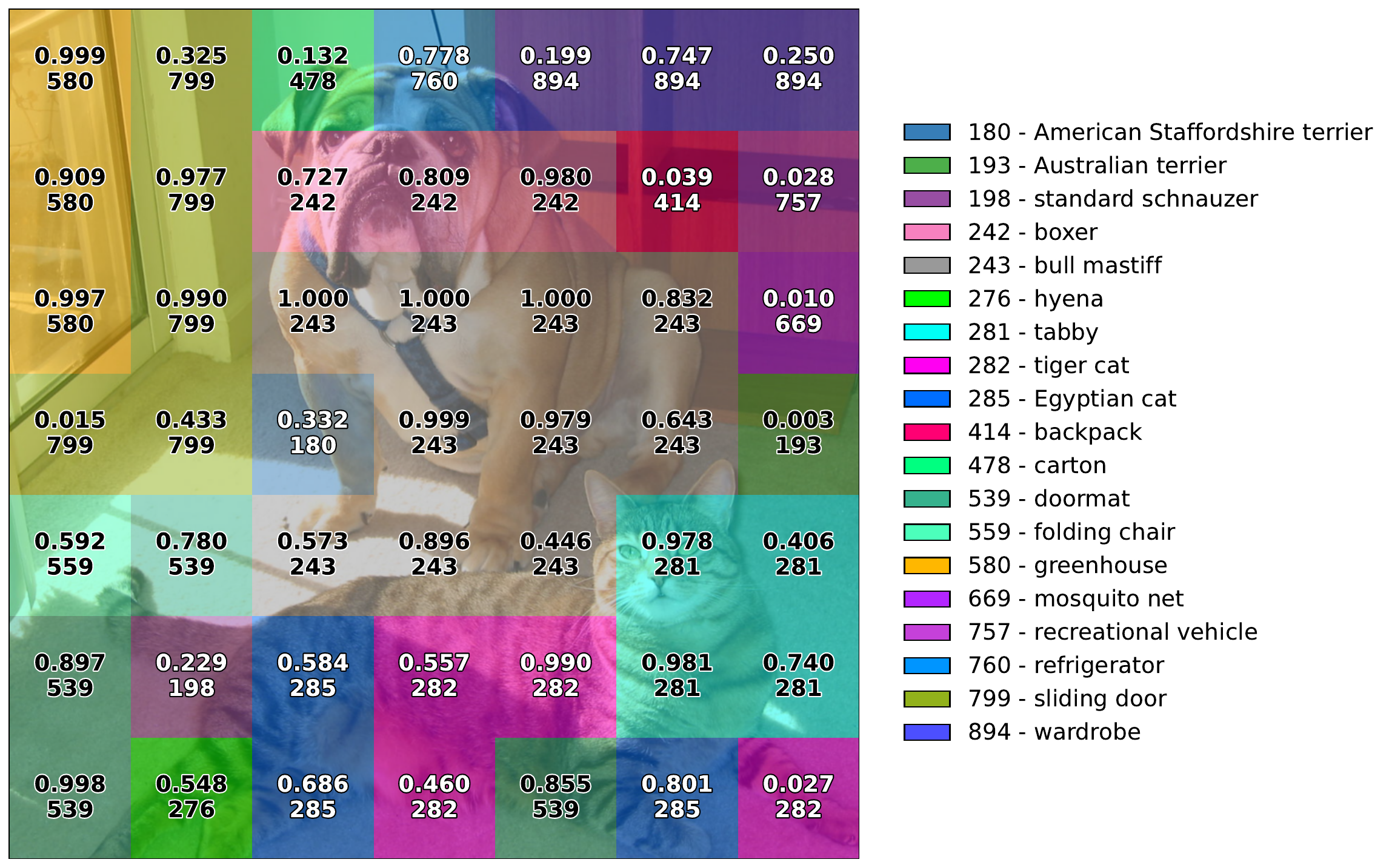}
    \caption{Original image}
    \label{fig:adv-original}
  \end{subfigure}
  \hfill
  \begin{subfigure}[b]{0.285\textwidth}
    \centering
    \includegraphics[width=\linewidth, height=4cm, keepaspectratio]{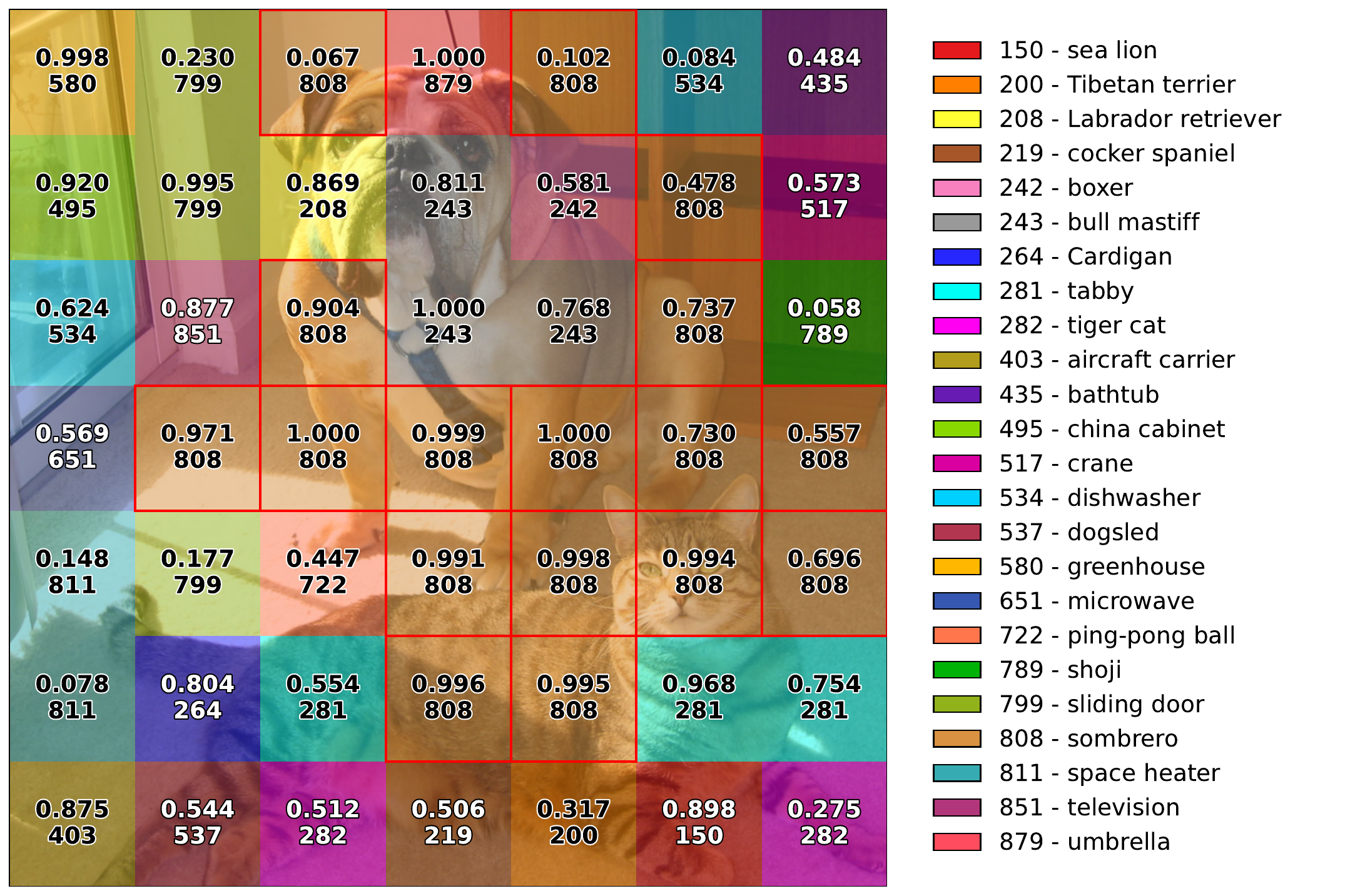}
    \caption{Perturbed image}
    \label{fig:adv-perturbed}
  \end{subfigure}
  \hfill
  \begin{subfigure}[b]{0.40\textwidth}
    \centering
    \includegraphics[width=\linewidth, height=4cm, keepaspectratio]{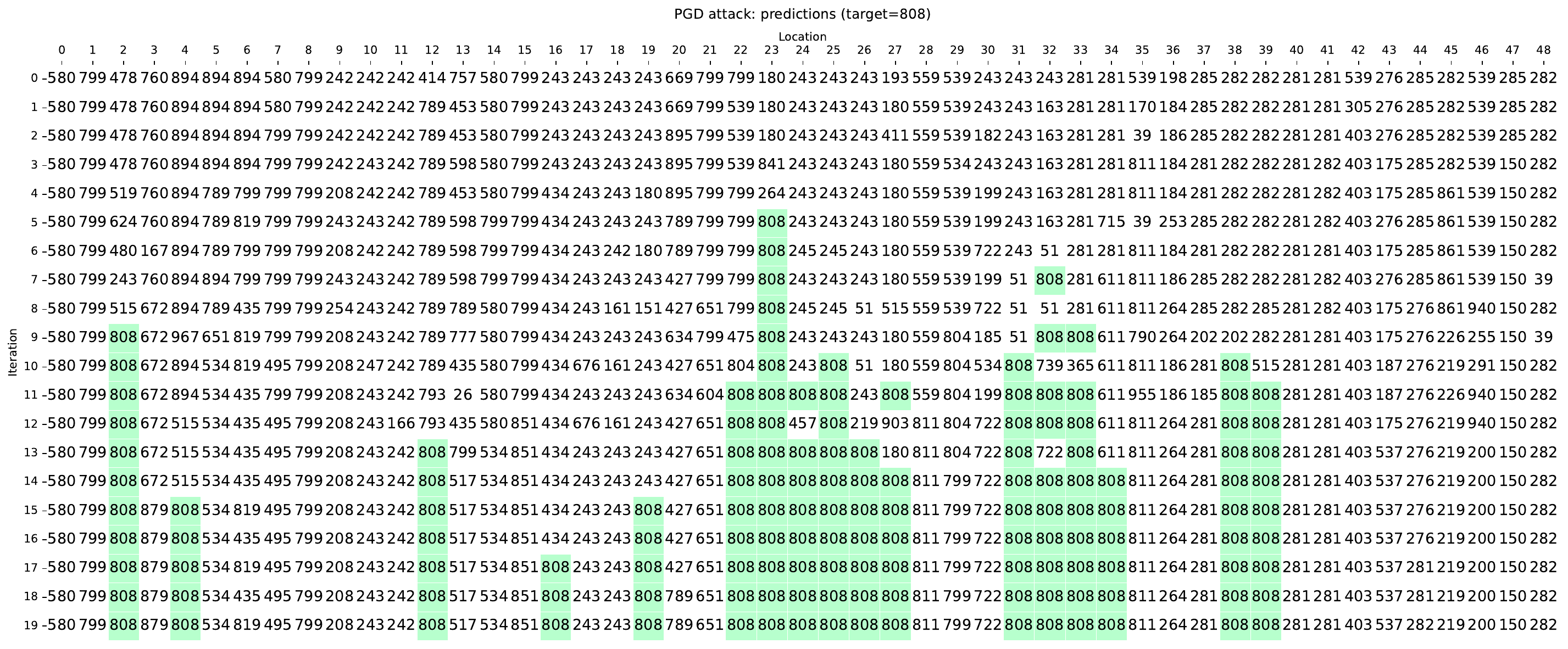}
    \caption{Per-location predictions across iterations}
    \label{fig:adv-iterations}
  \end{subfigure}
  \caption{Spatial decomposition of a targeted projected gradient descent (PGD) attack~\cite{madry2017towards}. The attack optimizes for the target class \texttt{sombrero} (ImageNet ID 808) over 20 iterations. (\subref{fig:adv-original}) and (\subref{fig:adv-perturbed}) display the original and perturbed images overlaid with patch-level top-1 predictions (class ID and softmax score), where target-class predictions are boxed in red. (\subref{fig:adv-iterations}) tracks the evolution of the patch-level top-1 predictions across all PGD iterations, revealing the localized spatial progression of the adversarial perturbation.}
  \label{fig:adversarial-attack}
\end{figure}

To illustrate this, we run a targeted projected gradient descent (PGD) attack~\cite{madry2017towards} on a standard pretrained classifier. Figure~\ref{fig:adversarial-attack} visualizes the spatial evolution of the attack toward an arbitrary target class (\texttt{sombrero}). Rather than uniformly shifting the global embedding, the iterative perturbation systematically flips specific spatial locations (Figure~\ref{fig:adversarial-attack}\subref{fig:adv-iterations}). Evaluating the spatial predictions on the final perturbed image (Figure~\ref{fig:adversarial-attack}\subref{fig:adv-perturbed}) reveals the mechanics of the failure: the attack succeeds by flipping a sufficient number of local patch predictions to the target class, which then dominates the correct local evidence during mean aggregation. This spatial diagnostic provides a granular view of model vulnerability, showing that adversarial susceptibility can be isolated and analyzed at the pre-aggregation level.

\begin{figure}[th]
  \centering
  \begin{subfigure}[b]{0.495\textwidth}
    \centering
    \includegraphics[width=\linewidth, keepaspectratio]{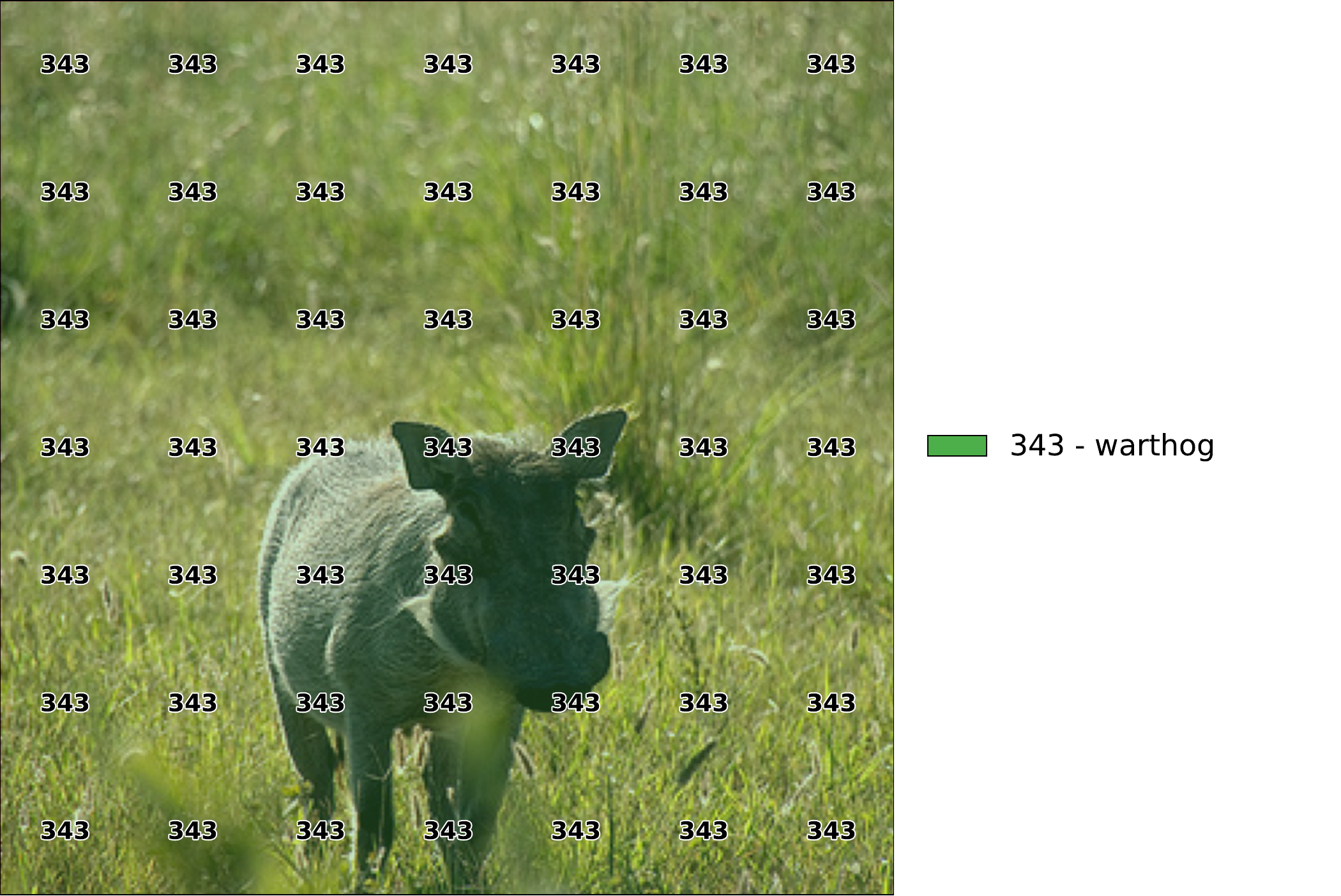}
    \caption{Predictions}
  \end{subfigure}
  \hfill
  \begin{subfigure}[b]{0.495\textwidth}
    \centering
    \includegraphics[width=\linewidth, keepaspectratio]{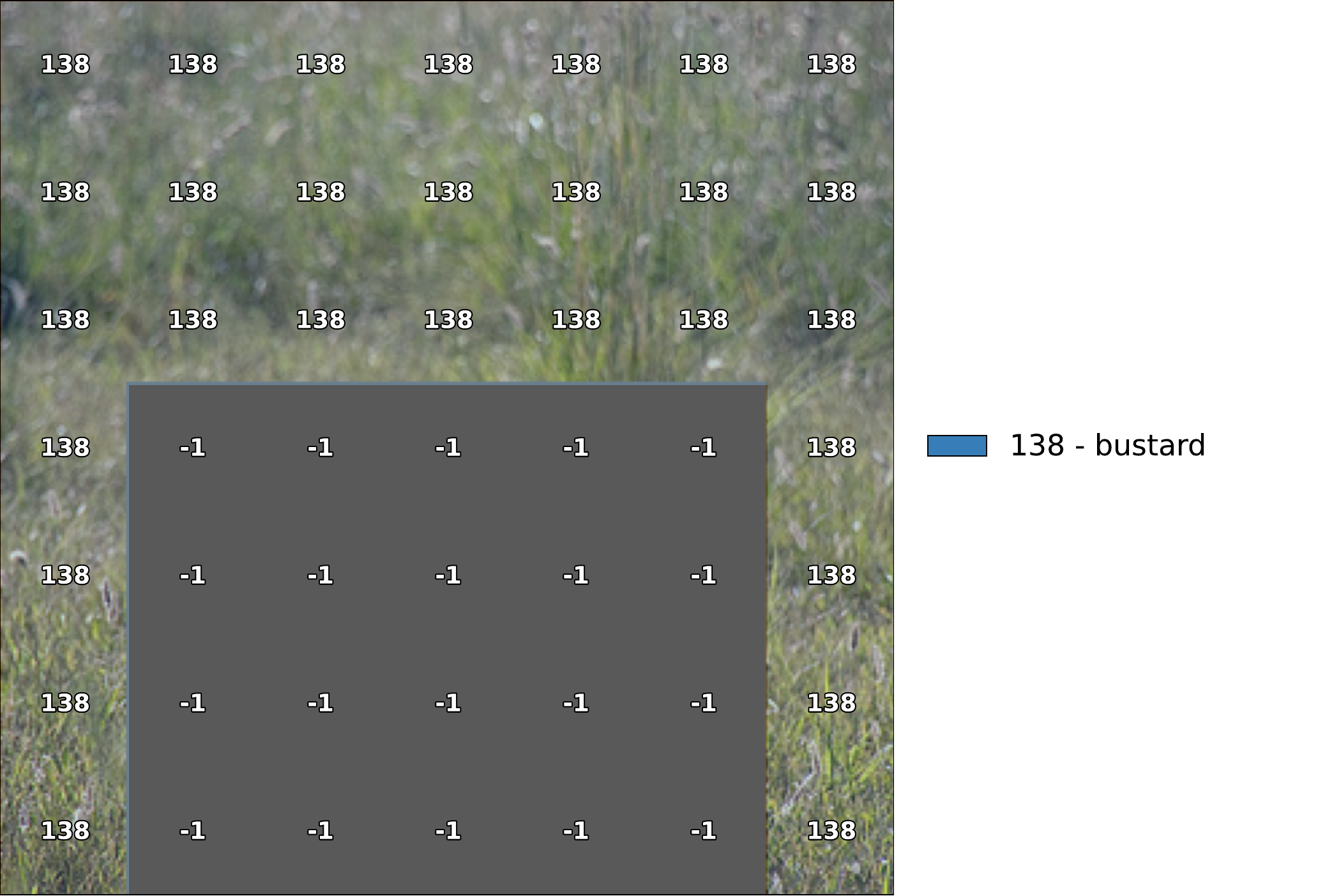}
    \caption{Occluded predictions}
  \end{subfigure}
  \caption{Context aggregation in ViT-based models. When the \texttt{warthog} is visible, background-token predictions can reflect object context through self-attention. When the object is occluded, no such evidence is available, and the per-token classifier produces arbitrary predictions on the unsupported background.}
  \label{fig:warthog-occlusion}
\end{figure}

%%%%%%%%%%%%%%%%%%%%%%%%%%%%%%%%%%%%%%%%%%%%%%%%%%%%%%%%%%%%%%%%%%%%%%%%%%%%%%%
%%%%%%%%%%%%%%%%%%%%%%%%%%%%%%%%%%%%%%%%%%%%%%%%%%%%%%%%%%%%%%%%%%%%%%%%%%%%%%%

\end{document}